\pdfoutput=1
\documentclass[11pt]{article}

\usepackage{acl}

\usepackage{times}
\usepackage{latexsym}
\usepackage{graphicx} % Required for inserting images
\usepackage{booktabs} % for pretty tables
\usepackage{lipsum} % for gibberish text
\usepackage{stfloats}
\usepackage{tabularx}
\usepackage{soul}
\usepackage{adjustbox}
\usepackage{appendix}
\usepackage{ctable}
\usepackage{amssymb}
\usepackage{mathtools, nccmath}
\usepackage{amsmath}
\usepackage{stmaryrd}  
\usepackage{caption}
\usepackage{subcaption}
\usepackage{multirow}
\usepackage{epstopdf}

\usepackage{paralist} %compactitem
\usepackage[utf8]{inputenc}
\usepackage{multirow}
\usepackage{xstring}
\usepackage{paralist} %compactitem
\usepackage[T1]{fontenc}
\usepackage[utf8]{inputenc}

\usepackage{microtype}

\usepackage{inconsolata}

\newcommand{\bl}[1]{{\color{black} #1}}

\title{Words of Warmth:\\ Trust and Sociability Norms for over 26k English Words}

\author{Saif M. Mohammad  \\
  National Research Council Canada \\
  {\tt saif.mohammad@nrc-cnrc.gc.ca} }
  
\begin{document}
\maketitle
\begin{abstract}

Social psychologists have shown that \textit{Warmth (W)} and  \textit{Competence (C)} are the primary dimensions along which we assess other people and groups. These dimensions impact various aspects of our lives from social competence and emotion regulation to success in the work place and how we view the world. More recent work has started to explore how these dimensions develop, why they have developed, and what they constitute. Of particular note, is the finding that warmth has two distinct components: Trust \textit{(T)} and \textit{Sociability (S)}. 
% Often, work on W and C makes uses of language in the form of responses from people to researcher questions in labs, despite known biases. Yet, work with every-day utterances and social media data is scarce.
% (trustworthyness (T)and sociability)
In this work,
we introduce \textit{Words of Warmth}, the first large-scale repository of manually derived word--warmth (as well as word--trust and word--sociability) associations for over 26k English words. We show that the associations are highly reliable.
% split-half reliability of 0.82 Spearman and 0.89 Pearson correlation. 
We use the lexicons to study the rate at which children acquire WCTS words with age. Finally, we show that the lexicon enables a wide variety of bias and stereotype research 
% in psychology, NLP, public health, and social sciences 
through case studies on various target entities.
Words of Warmth is freely available at:\\[1pt] 
\url{http://saifmohammad.com/warmth.html}
% \url{http://saifmohammad.com/Words of Warmth.html}

%, and digital humanities.
\end{abstract}

\section{Introduction}

\hspace*{1.4cm} \textit{Who goes there: friend or foe?}\\
\noindent \bl{This is a question human beings have asked from
the earliest of times to the present day.}
% This is a question asked not just by sentries from an age long gone but by all human beings in general---from the earliest of times to the present, in one form or another. %\\[3pt] 
  %\noindent{\textbf{Warmth.}} 
  A large body of social psychology research has shown that \textit{warmth (W)}  (friendliness, trustworthiness, and sociability) and \textit{competence (C)} (ability, power, dominance, and assertiveness) are core dimensions of social cognition and stereotypes \cite{fiske2002,bodenhausen2012social,fiske2018stereotype,abele2016facets,koch2024validating}.
  That is, human beings quickly and subconsciously judge (assess) other people, groups of people, and even their own selves along the dimensions of warmth and competence---likely because of evolutionary pressures \cite{macdonald1992warmth,eisenbruch2022warmth}. Assessing W and C was central to early human survival (e.g., to anticipate whether someone will help them build useful things or whether they might steal their resources).

%  Warmth is an indicator of intent: friendly or unfriendly. 
% and it continues to be pivotal in modern-day life for success in a social work environment, for success in mate selection (dating), and for success in managing inter-personal relationships in general. Warmth also plays an important role in how we view people and social groups, such as the leaders of a country, immigrants, elderly, and homeless. %\\[3pt] 
% Thus human beings ascertain (perceived) warmth very quickly from any new social interactions often without even consciously realizing it \cite{}.\\[3pt] 
%
% \noindent{\textbf{Competence.}} 
% The other primary dimension of social cognition is competence (aka dominance). It refers to the dimension of capability, agency, powerfulness, and strength. Similar to warmth, the ability to assess competence has strong evolutionary benefits (e.g., to anticipate whether a friend can help build great things or an enemy has the ability to cause great harm); and it is continues to be important in modern life.\\[3pt] 
%
% \noindent{\textbf{Why Warmth--Competence Assessments Matter.}}
% \noindent{\textbf{Warmth--Competence Quadrants, Ingroup vs.\@ Outgroup, Stereotypes, and Emotions.}} 

The dimensions of W and C have been shown to have substantial implications on a wide variety of facets, including: interpersonal status \cite{swencionis2017warmth}, social class \cite{durante2017social}, self-beliefs \cite{wojciszke2009two}, political perception \cite{fiske2014never}, child development \cite{ROUSSOS2016133},  cultural analyses \cite{fiske2016stereotype}, as well as
professional and organizational outcomes, such as hiring, employee evaluation, and allocation of tasks and resources \cite{cuddy2011dynamics}. 

W and C are considered to be orthogonal and together they create four quadrants: high W and high C, low W and high C, low W and low C, high W and low C. 
% One may be perceived as warm and competent, warm and incompetent, cold and competent, or cold and incompetent.
The \textit{Stereotype Content Model} \cite{fiske2002} argues that how we perceive others is influenced by whether they are considered to be members of the \textit{ingroup} (the same country, political affiliation, language, etc.) or \textit{outgroup} (a different country, political affiliation, language, etc.).
Members of the ingroup are generally considered to be high W and high C, whereas members of the outgroup tend to be perceived consistent with the other quadrants. For example, it has been shown that the \textit{stereotypical} view towards members of one's own social class is that they are high W and C, whereas the poor and homeless are perceived as low W (cold) and low C (incompetent), the elderly are perceived as high W and low C, and accountants and business people are perceived as low W and high C \cite{fiske2018stereotype}. 

These perceptions and stereotypes (influenced by ingroup and outgroup memberships) evoke different emotions and behaviour. For example, a positive event associated with someone in our ingroup (considered warm and competent) evokes pride, whereas a   positive event associated with someone in our outgroup (e.g., considered cold and competent) evokes envy. \bl{Thus determining W and C perceptions is tremendously valuable in understanding: why people act the way they do;  what is driving the discourse in complex social interactions such as discussions about climate change;  how different social groups (e.g., immigrants, disabled people, and elderly) are viewed by different groups; and whose view of the world is centered.}

% \noindent{\textbf{Trustworthiness and Sociability.}} 
More recently, psychologists have shown that warmth should be modeled in terms of two separate dimensions: \textit{Trust (T)} and \textit{Sociability (S)} \cite{abele2016facets,koch2024validating}. T is the dimension of trust, morality, goodness, sincerity, and integrity, whereas
S is the dimension of sociableness, friendliness, gregariousness, and conviviality.
(As shorthand, we will refer to any set of dimensions by simply their letters: WTS for warmth, trust and sociability, WCTS for all four, etc.)
% These are clearly different dimensions, but one can also see why they go together: e.g., we tend to be more friendly towards people we find to be moral. Nonetheless, in certain situations, it is useful to do a finer-grained analysis of trustworthyness and sociability separately, rather than a combined analysis of warmth. For instance, to understand social perceptions of morality and warmth of society towards unwed mothers, the poor, homosexual people, etc. \\[3pt]
%
%\noindent{\textbf{Language.}} 

W and C are perceived through various modalities, including: facial expressions, body language, one's actions, what one says, how they say it (words used, tone, etc.). Language is of particular interest as it is a direct and vastly expressive medium.
% for communicating one's intentions. 
Often, work on W and C makes use of language in the form of responses to researcher questions in labs. However, a notable issue is that people can be reluctant to explicitly divulge their stereotypes towards certain target groups \cite{nosek2005understanding,maina2018decade,hilton1996stereotypes}. 
Thus, work with every-day utterances and social media data is attractive as a complement to traditional approaches.
Further, the words one uses can often communicate W and C through associations (connotations), and can reveal perceptions and stereotypes 
% that one may not be consciously aware.
\bl{(even if the speaker is not consciously aware of it).}

\bl{Large manually compiled repositories of word--competence norms exist for English: e.g., the NRC VAD Lexicon $\sim$ 20k words---used widely for sentiment analysis research. However, existing lexicons for W are much smaller: e.g., \citet{nicolas2021comprehensive} manually compiled a set of 341 words.}\\[3pt]
\noindent{\textbf{Our Work.}} We \bl{compiled} sociability and trust \bl{association} norms for over 26k English words. 
The lexicons were created by crowdsourcing and employing a slate of quality control measures.
We show that the resulting association scores have high reliability (repeating the annotations leads to very similar scores and rankings).
We created a third combined lexicon for warmth by taking the union of the entries for the trust and sociability lexicons. Together, we refer to \bl{the} set of three \bl{lexicons} as the \textit{Words of Warmth Lexicons}.%\\[3pt]

The three lexicons enable a wide variety of research and applications. Notably:
% social science, psychology, and NLP research in the areas of stereotypes, social behaviour, public messaging, social competence, emotion regulation, etc. 

% \begin{compactenum}
    
% \item 
\noindent \textit{In Psychology and Social Cognition}
\begin{compactitem}
\item 
 What kind of trust assessments do children develop first? And what kinds are developed later? (Trust can be of different kinds: care-based, character-based, consistency-based, etc.) Similarly for sociability.
\item 
 What are the mechanisms underpinning the development of WCST assessment capabilities in children?
How does exposure to different conditions impact these capabilities?
\item 
 How different are the WCST capabilities of people in different cultures?
\item 
What role do differences in language play in the development of WCST capabilities? 
%(This will require lexicons created in other languages as well; but having one for English is a start.)
% \item What causes changes in how we perceive WCST?
\end{compactitem}
% \item 
\textit{In Computational Social Science, NLP}
\begin{compactitem}
\item
The lexicons can be used to study public discourse on topics of interest. For example, how are the levels of warmth, competence, trust, and sociability in online discussions about climate change or vaccines changing with time; how do these levels vary for different stakeholders?; what sub-aspects of climate change (or vaccines or any topic of interest) evoke the lowest amounts of warmth, competence, trust, and sociability? etc.
\item 
 How has the perceived WCST of a chosen target of interest (say government, banks, immigrants, etc.) changed over the last 100 years? 
\end{compactitem}
% \item 
\textit{In HCI and NLP}
\begin{compactitem}
\item 
Understanding perceptions of WCST of people towards artificial agents.
% (e.g., see papers: Humans perceive warmth and competence in artificial intelligence; Warmth and competence in human-agent cooperation).
\end{compactitem}
%\item 
\textit{In Digital Humanities and NLP}
\begin{compactitem}
\item 
 What role do warmth, trust, sociability, and competence play in developing compelling characters and story arcs? How does this vary by genre and culture?
\end{compactitem}
% \item 
\textit{In Commerce}
\begin{compactitem}
\item 
 Tracking warmth, trust, sociability, and competence towards one’s product on social media. This can help understand product branding, tracking user satisfaction, and taking the appropriate remedial actions for product improvement and public-facing communications.
\item 
 Understanding how perceptions of warmth and competence of one’s product impact customer behavior. 
% (e.g. papers: ] Warmth trumps competence? Uncovering the influence of multimodal AI anthropomorphic interaction experience on intelligent service evaluation; The effect of virtual anchor appearance on purchase intention: a perceived warmth and competence perspective)
\end{compactitem}

%\end{compactenum}

% We list some of these broad directions below. 
In the second half of this paper, we use the lexicons to explore:
\setdefaultleftmargin{1em}{}{}{}{.5em}{.5em}
\begin{compactenum}
% \begin{itemize}
%    \item How are warmth, trust, and sociability related to various other dimensions and categories of emotions such as valence, arousal, anger, and sadness? Exploring this question sheds light on how perceptions of warmth co-occur with other emotions, and therefore what implications these have on our behaviour.  
    \item At what rate do children acquire WCTS words? And how do these change with age? This sheds light on how social cognition develops and on the relative importance of the two dimensions. (\ref{sec:children}) % (overall, and across age).
    \item How do we use W and C words in social media, especially when mentioning various social groups? This sheds light on how our perceptions of W and C towards various social groups manifests in public discourse.
% \end{itemize}
\end{compactenum}
\noindent We make all of the lexicons and code freely available for research.\footnote{\url{http://saifmohammad.com/warmth.html}}

\section{Related Work}

Despite the considerable importance of warmth and competence in social cognition and behaviour (as discussed in the Introduction), there is much we do not know about
how these dimensions develop; how children assess W and C of those around them;
% to determine who to trust and who to learn from; 
and which dimension is of greater significance.

% and how warmth and competence of caregivers impacts child development, social competence, and emotion regulation.  
Some research argues that warmth is the primary component of valence, which in turn is evolutionarily central to the approach--avoid response, and so  assessment of warmth emerges earlier than competence \cite{cuddy2007bias}.
This is the \textit{primacy of valence} hypothesis.
The view that, in children, competence emerges earlier than warmth, is known as the \textit{primacy of competence hypothesis} \cite{ROUSSOS2016133}. In support of this hypothesis are some studies that show that infants (even as young as 6 to 8 months) assess competence levels and show more trust in those they think are more competent \cite{koenig2003infants,tummeltshammer2014infants}. 
% https://www.sciencedirect.com/science/article/pii/S0022096515002027?casa_token=1T6kcYSiImIAAAAA:82fX2PccaEDre3AeYj4wuN8vrRKeH_V9OCHChXH-xCSbMzSNv9AkqmKYTeMYvO15RRmuc7D9xws
Finally, there are studies showing how warmth towards the child from the caregiver has tremendous positive benefits for the child, arguably again showing that warmth is more important for children than competence. For example, 
\citet{altschul2016hugs} show that spanking by the caregiver predicted increases in child aggression. In contrast, caregiver warmth (much more than spanking) predicted social competence. 
%brb% \cite{macdonald1992warmth} argue that warmth evolved to facilitate cohesive family ties and the acceptance of adult values, and central to personality trait development. 

Since words act as the principal carriers of meaning, and many words connotate W and C, large lexicons of W and C associations can be powerful resources for understanding questions such as those discussed above. There exist many lexicons for competence (aka dominance), such as the 
 \newcite{warriner2013norms} and \newcite{mohammad-2018-obtaining} for English;  \newcite{moors2013norms} for Dutch, and  \newcite{vo2009berlin} for German. % and by \newcite{redondo2007spanish} for Spanish).
The largest among these is the NRC VAD lexicon \cite{mohammad-2018-obtaining,vad-v2}: version 1 has entries for over 20,000 English words, and version 2 for over 44,000 unigrams and 10,000 bigrams (two-word sequences).
However, manually compiled lexicons for word--warmth associations are much smaller. Most notably, \citet{nicolas2021comprehensive} manually compiled a set of 341 words. They also expanded this lexicon automatically using WordNet synonyms and word embeddings. However, even near synonyms and distributionally close term pairs can convey very different WST associations; for example, slip vs.\@ fault and skinny vs.\@ slender. In fact, \citet{fraser-etal-2024-stereotype} found that the automatically expanded lexicon was not effective in capturing W and C. 

In response to the tremendous upsurge of social media content, generative AI, polarization, and rising misinformation, we have also seen growing interest in tackling  stereotypes and bias research in NLP. 
Influential early work explored race and gender bias in word embeddings and automatic systems  \cite{caliskan2017semantics,thelwall2018gender,kiritchenko-mohammad-2018-examining,tan2019assessing,blodgett2020language}. 
A considerable amount of recent research explores bias and stereotypes in generative AI \cite{kotek2023gender,zhou2024bias,baines2024playgrounds}. 
Yet, a growing area of interest is work on exploring bias and stereotypes in large amounts of social media data from a computational social science perspective \cite{sanchez2021you,ariza2022overview,bosco2023detecting,fraser-etal-2024-stereotype,schmeisser2024stereohoax}.

% As part of the \textit{Words of Warmth Project}, we created large lexicons 
% of word--trust, word--sociability, and word--warmth associations 
% trust, sociability, and warmth. %with a focus on reliable annotations.
% \footnote{In follow on work, we also compiled anxiety association norms for over 10K English phrases \cite{ww-2}.}
% (larger even than existing lexicons for VAD and categorical emotions such as anger and fear). 
The WST lexicons we created are useful in studying the core dimensions of social cognition, how they develop in children, how they impact our traits, and how they shape our views and stereotypes.

% We then use the lexicons to %shed light on the importance of these dimensions through 
% quantify the rate at which children acquire words associated with these dimensions. Finally, we show how the lexicons can be used to assess stereotypes in social media data. 

\section{Obtaining Human Ratings for Trust, Sociability, and Warmth}

% The keys steps in creating the anxiety dataset were: 
% \begin{compactenum}
% \item selecting the terms to be annotated 
% \item developing the questionnaire
% \item developing measures for quality control (QC)
% \item annotating terms on a crowdsource platform
% \item discarding data from outlier annotators (QC)
% \item aggregating data from multiple annotators to determine the anxiety association scores
% % for each of the terms. 
% \end{compactenum}
We describe the main steps below.\\[3pt]
\noindent {\bf 1. Term Selection.}
% Our goal was to annotate a large number of common English terms. 
% the NRC Emotion Lexicon \cite{MohammadT13,MohammadT10}, 
% the terms from the Prevalence dataset \cite{brysbaert2019word} 
We wanted to include a large set of common English words. Further, we wanted to especially include terms with emotion associations (as opposed to lots of emotionally neutral terms).  
Thus, we chose the NRC VAD Lexicon \cite{mohammad-2018-obtaining} as the source of terms. Version 2 includes $\sim$44k unigrams 
%brb and 10k bigrams 
annotated for valence, arousal, and dominance. The valence (or sentiment) scores go from -1 (maximum negativeness to +1 (maximum positiveness). Scores between -0.33 and +0.33 correspond to neutral valence. After manual examination of the valence scores, we chose to exclude terms with a valence score between -0.2 and +0.2 (keeping all of the non-neutral terms as well as some neutral terms).
This resulted in a set of % 31,535 terms in total (26,188 unigrams and 5,347 bigrams).
26,188 unigrams.\\[3pt]
% (specifically, words marked as being known to over 70\% of the respondents). 
% Summary details of each of these is provided in the Appendix.
\noindent{\bf 2. Trust and Sociability Questionnaires.}
The questionnaires used to annotate the data 
 were developed  after several rounds of 
 pilot annotations. 
 Detailed directions, including notes directing respondents to consider predominant word sense (in case the word is ambiguous) and example questions (with suitable responses) were provided. (See Appendix.)
 The primary instruction and the questions presented to annotators are shown below.\\[-17pt]
% The annotation questions and the instructions for the annotators are shown in a supplementary file. 

{
\noindent\makebox[\linewidth]{\rule{0.48\textwidth}{0.4pt}}\\% [-8pt]
{ \small
% Summary Instructions
\noindent Consider trustworthiness to be a broad category that includes:\\
\indent \textit{trustworthy, honesty, fairness, dependability, reliability, \\ 
\indent morality, virtuousness, sincerity, honorableness, etc.}\\
Consider untrustworthiness to be a category that includes:\\
\indent \textit{unfairness, dishonesty, untrustworthiness, dubiousness, \\ 
\indent immorality, sinfulness, insincerity, dishonorableness, etc.}
% nonchalant, uninterested, 
% If you do not know the meaning of a word or are unsure, you can look it up in a dictionary (e.g., the Merriam Webster) or on the internet.

% Quality Control

% Some questions have pre-determined correct answers. If you mark these questions incorrectly, we will give you immediate feedback in a pop-up box. An occasional misanswer is okay. However, if the rate of misanswering is high (e.g., >20\%), then all of one's HITs may be rejected

%bb% \noindent Select the options that most English speakers will agree with.\\[4pt]
\noindent Q1.  <term> is often associated with feeling:\\[-1pt]
\indent 3: very trustworthy \hspace{11mm} -1: slightly untrustworthy\\[-1pt]
\indent 2: moderately trustworthy \hspace{3mm}  -2: moderately untrustworthy\\[-1pt]
\indent 1: slightly trustworthy \hspace{7mm}  -3: very untrustworthy\\[-1pt]
\indent 0: not associated with being trustworthy or untrustworthy \\[-8pt]
% \indent -1: slightly calm\\
% \indent -2: moderately calm\\
% \indent -3: very calm\\[-8pt]
}
\noindent\makebox[\linewidth]{\rule{0.48\textwidth}{0.4pt}}\\[-24pt]

}

{
\noindent\makebox[\linewidth]{\rule{0.48\textwidth}{0.4pt}}\\% [-8pt]
{ \small
% Summary Instructions
\noindent Consider social warmth to be a broad category that includes:\\
\indent \textit{warmness, sociableness, generosity, helpfulness, \\ 
\indent tolerance, understanding, thoughtfulness, etc.}\\
Consider social coldness to be a broad category that includes:\\
\indent \textit{coldness, antisocialness, ungenerosity, unhelpfulness, \\ 
\indent intolerance, indifferent, thoughtlessness, etc.}
% nonchalant, uninterested, 
% If you do not know the meaning of a word or are unsure, you can look it up in a dictionary (e.g., the Merriam Webster) or on the internet.

% Quality Control

% Some questions have pre-determined correct answers. If you mark these questions incorrectly, we will give you immediate feedback in a pop-up box. An occasional misanswer is okay. However, if the rate of misanswering is high (e.g., >20\%), then all of one's HITs may be rejected

%bb% \noindent Select the options that most English speakers will agree with.\\[4pt]
\noindent Q1.  <term> is often associated with feeling:\\[-1pt]
\indent 3: very sociable \hspace{14mm} -1: slightly unsociable\\[-1pt]
\indent 2: moderately sociable \hspace{5mm}  -2: moderately unsociable\\[-1pt]
\indent 1: slightly sociable \hspace{10mm}  -3: very unsociable\\[-1pt]
\indent 0: not associated with being sociable or unsociable \\[-8pt]
% \indent -1: slightly calm\\
% \indent -2: moderately calm\\
% \indent -3: very calm\\[-8pt]
}
\noindent\makebox[\linewidth]{\rule{0.48\textwidth}{0.4pt}}\\[-12pt]

}
% \noindent The full questionnaire will be made available on the project webpage. 

\noindent{\bf 3. Quality Control Measures.}
%brb author no s below
About 2\% of the data was annotated beforehand by the authors and interspersed with the rest. These questions are referred to as \textit{gold} (aka \textit{control}) questions. 
% During crowd annotation, 
% We interspersed the gold questions with the other questions.
% and the annotator is not aware which is which. However, 
Half of the gold questions were used to provide immediate feedback to the annotator (in the form of a pop-up on the screen) in case they mark them incorrectly. We refer to these as \textit{popup gold}. This helps prevent the situation where one annotates a large number of instances without realizing that they are doing so incorrectly. 
It is possible, 
that some annotators share answers to gold questions with each other (despite this being against the terms of annotation). 
% The gold questions also served as examples to guide the annotators.
Thus, the other half of the gold questions were also separately used to track how well an annotator was doing the task, but for these gold questions no popup was displayed in case of errors. 
We refer to these as 
\textit{no-popup gold}.\\[3pt]
% If a crowd worker answered a gold question incorrectly, then they were immediately notified.
% \sm{the annotation was discarded, and an additional annotation was requested from a different annotator}. 
%
\noindent{\bf 4. Crowdsourcing.} 
We setup the annotation tasks on the crowdsourcing platform, {\it Mechanical Turk}.
 In the task settings, we specified that we needed annotations from nine people for each word.
 (Since we got some additional funding later, three more annotations per word were obtained for trust.) 
We obtained annotations from native speakers of English residing around the world. Annotators were free to provide responses to as many terms as they wished. 
The annotation task was approved by 
\bl{the National Research Council Canada's Institutional Review Board}. 
% our institution's review board (IRB number anonymized).
% , which reviewed the proposed methods to ensure that they were ethical.
% We used CrowdFlower's gold annotations scheme for quality control, wherein 

\noindent {\it Demographics:} \bl{Hundreds of annotators participated in each of the annotation tasks. About 69\% of the respondents % who annotated the words 
live in USA. The rest were from India, United Kingdom, and Canada.} 
The average age of the respondents was 39.2 years. \bl{Among those that provided a response to the gender question: about 48\% entered female, 52\% said male, and no one marked themselves as nonbinary (or provided any other response).} %brb% \footnote{Respondents were shown optional text boxes to disclose their demographic information as they choose; especially important for social constructs such as gender, in order to give agency to the respondents and to avoid binary language.}\\[-13pt]

\noindent{\bf 5. Filtering.} % and Re-annotation} %  as part of Quality Control} 
If an annotator's accuracy on the gold questions (popup or non-popup) fell below 80\%, then they were refused further annotation, 
 and all of their annotations were discarded (despite being paid for).
% We then obtained fresh annotations for those terms from other annotators.
% This served as a mechanism to avoid malicious and random annotations.
  % However, because of the way the gold questions work in CrowdFlower, they were annotated by more than six people. 
%  Both the minimum and the median number of annotations per item was 10. 
 See Table~\ref{tab:ann} for summary statistics.\\[-13pt]

\begin{table*}[t!]
\centering
% \begin{center}
%\vspace*{-4mm}
\small{
\begin{tabular}{lrrrrrrrr}
\hline 
%       & \multicolumn{2}{c}{\bf \#instances from} &\\
%				 & Tweets-2017 & Tweets-2018 &\\

{\bf Dataset} 	& \bf \#words	&\bf Annotators    &\bf \#Annotations &\bf MAI  &\bf SHR ($\rho$)		 &\bf SHR ($r$) \\\hline
sociability 		& 26,123 &US, India, UK, Canada   & 205,475  & 7.9 	& 0.965 &0.969\\
trust 		& 26,185 &US, India, UK, Canada   & 299,365  & 11.4 	& 0.943 &0.957\\
 warmth 		& 26,085 &US, India, UK, Canada   &229,580 &8.8	&0.965 &0.974\\
% bigrams 		& 10,000 &worldwide  & 1,020 & 10 	& 243,295\\
%\hline
%\bf Total  &54,000 		 	& & & &  \bf 778,085\\
 \hline
\end{tabular}
}
%\vspace*{-1mm}
\caption{\label{tab:ann} {A summary of the Words of Warmth annotations.  MAI = mean  annotations per word. SHR, measured through both Spearman rank and Pearson's correlations (last two columns), indicate high reliability.}
%brb The warmth lexicon created from the union of the sociability and trust lexicons has 26,085 words.
}
 \vspace*{2mm}
% \end{center}
\end{table*}

% warmth-unigrams-lexicon.twb
 \begin{figure*}[t]
	     \centering
	     \includegraphics[width=\textwidth]{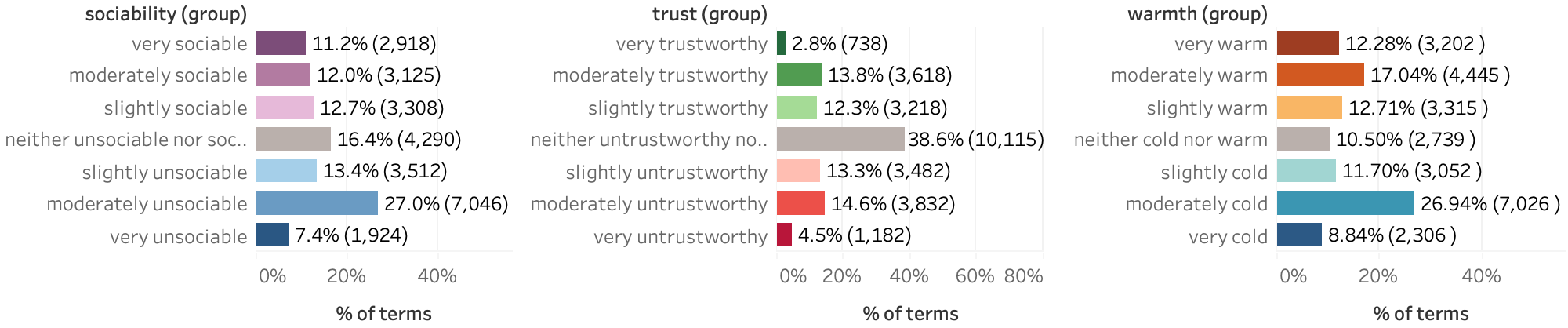}
	    \vspace*{-5mm}
      \caption{Distribution of terms in Words of Warmth: percentage and number of terms associated with each class.}
	     \label{fig:Words of Warmth-distrib}
\vspace*{-5mm}
	 \end{figure*}

\noindent{\bf 6. Aggregation.} 
Every response was mapped to an integer from -3 (very untrustworthy/unsociable) to 3 (very trustworthy/sociable). % as follows: 
% \begin{compactitem}
%     \item high anxiety: 3
%     \item moderate anxiety: 2
%     \item slight anxiety: 1
%     \item neither anxiety nor calmness: 0
%     \item slight calmness: -1
%     \item moderate calmness: -2
%     \item high calmness: -3
% \end{compactitem}
The final  score for each term is simply the average score it received from each of the annotators.
% The scores were then linearly transformed to the interval: -1 (highest calmness) 
% to 1 (highest anxiety).
% Since degree of emotion is a unipolar scale, 
% We linearly transform the -1 to 1 scores to scores in the range 0 (least emotion intensity) to 1 (the most emotion intensity).
We also created a categorical version of the sociability (S) lexicon by labeling all words that got an average S score $\geq 2.5$ as \textit{high S}, $\geq 1.5$ and $< 2.5$ as \textit{moderate S}, $\geq 0.5$ and $< 1.5$ as \textit{slight S},
$> -0.5$ and $< 0.5$ as \textit{neither sociable nor unsociability}, and so on. The categorical version of the trust (T) lexicon was created similarly.

% Finally, we created a warmth (W) lexicon by taking the union of the entries for T and S. 
% Why take the union (or’ing) of the S and T scores?
\bl{Social cognition research (Abele et al. 2016, Koch et al., 2024) points to how we perceive a person (or group) to be warm because we associate them with kindness, honesty, gregariousness, thoughtfulness, or some other quality associated with warmth. It is not required that one is both kind and gregarious or both honest and gregarious, etc. Thus, we 
created a warmth (W) lexicon by taking the union/or'ing of the entries for T and S.\footnote{That said, we also make the individual S and T scores available. So one can easily take the mean or some other function if that is more suitable for their particular application.}}
% if it is more useful to employ such a metric. One situation where being more conservative and taking the mean or harmonic mean may be more desirable is perhaps in the case of highly ambiguous words where one sense of the word is high T and a different sense is low S, and taking the mean effectively means a score close to 0 (and thus not listening to this word at all). However, such words are quite rare (<0.1\%). Overall, we believe in most applications it is more useful to take warmth to be the union as described in the paper. We will add the above discussion to the paper.

For a given word \textit{x}, 
if the absolute value of the T score for \textit{x} is greater than the absolute value of the S score for \textit{x}, then the W score for \textit{x} is taken to be the T score.
If the absolute value of the S score for \textit{x} is greater than the absolute value of the T score for \textit{x}, then the W score for \textit{x} is taken to be the S score. If the two scores are the same, then the same score is taken as the W score.
Thus, for example, for the word \textit{uplifting} with an S score of 3 and a T score 0.67, the W score is 3; \textit{birdbrain} with an S core of -1.71 and a T score of -2.62 gets a W score of -2.62. Figure \ref{fig:WTS} in the Appendix shows a scatter plot of words on the T--S space, colour coded as per their W score.
%% -- the warmth score (w) = max (absolute(trust score (w)), absolute(socibility(w)) 

We refer to the
list of words
along with their 
% real-valued final 
scores and categorical labels for WST as the {\it Words of Warmth Lexicons}. 
(Table \ref{tab:examples} in the Appendix shows example entries.)
\bl{Since warmth analyses are often done along with competence (aka dominance) analyses, we also include in the lexicon the competence scores for the terms (taken from the NRC VAD Lexicon v2 \cite{vad-v2}). Thus we also refer to this suite of lexicons as the \textit{Warmth--Competence Lexicons}, or the \textit{WCST Lexicons}.}
% with the highest and lowest scores.

Figure \ref{fig:Words of Warmth-distrib} shows the distribution of the different classes. 
\bl{As expected, most entries for trust are associated with neither trustworthyness nor untrustworthyness (38.6\%), but it is worth noting that 28.9\% of the words are associated with trustworthyness (to some degree) and 32.4\% of the words are associated with untrustworthyness (to some degree).
The pattern is different for sociability, wherein, a large number of inanimate objects are seen as moderately unsociable (the most frequent category). In the warmth lexicon, 10.5\% of the entries are marked as neither cold nor warm, whereas 42\% have some association with warmness and 47.5\% have some association with coldness.
(Figure \ref{fig:WbreakTS} in the Appendix shows a further break down of the percentage of terms in each of the warmth classes into the percentage of entries obtained from the trust lexicon and the percentage of entries obtained from the sociability lexicon.)}

\section{Reliability of the Annotations} 
A useful measure of quality is the reproducibility of the end result---repeated independent manual
annotations from multiple respondents should result in similar  scores.
To assess this reproducibility, we calculate
average {\it split-half reliability (SHR)} over 1000 trials.\footnote{SHR is a common way to determine reliability of responses to generate scores on an ordinal scale %in the fields of psychology and psycholinguistics 
\cite{weir2005quantifying}.} 
% SHR a commonly used approach to
% determine consistency in psychological studies, that we employ as follows.   
All annotations for an
item are randomly split into two halves. Two separate sets of scores are aggregated, just as described in Section 3 (bullet 6), from the two halves. 
% but independently from the two SHR halves.  
% Then the correlation between the two sets of scores is calculated. 
Then we determine how close the two sets of scores are (using a metric of correlation). % using Spearman Rank Correlation. 
This is repeated 1000 times and the correlations are averaged.
The last two columns in Table~\ref{tab:ann} show the results (split half-reliabilities). Spearman rank and Pearson correlation scores of around 
%0.96  for S and over 0.94 for T 
0.95 indicate very high reliability of the real-valued scores obtained from the annotations.\footnote{For reference, if the annotations were random, then repeat annotations would have led to an SHR of 0. Perfectly consistent repeated annotations lead to an SHR of 1. Also, similar past work on word--anxiety associations had SHR scores in the 0.8s \cite{mohammad-2024-worrywords}.} 
%  Past emotion lexicon work such as the NRC Emotion Intensity Lexicon \cite{} report SHRs in the low 0.9s with 32 annotations per item.)

% AoA-col.twb
     \begin{figure*}[t]
	     \centering
	     \includegraphics[width=\textwidth]{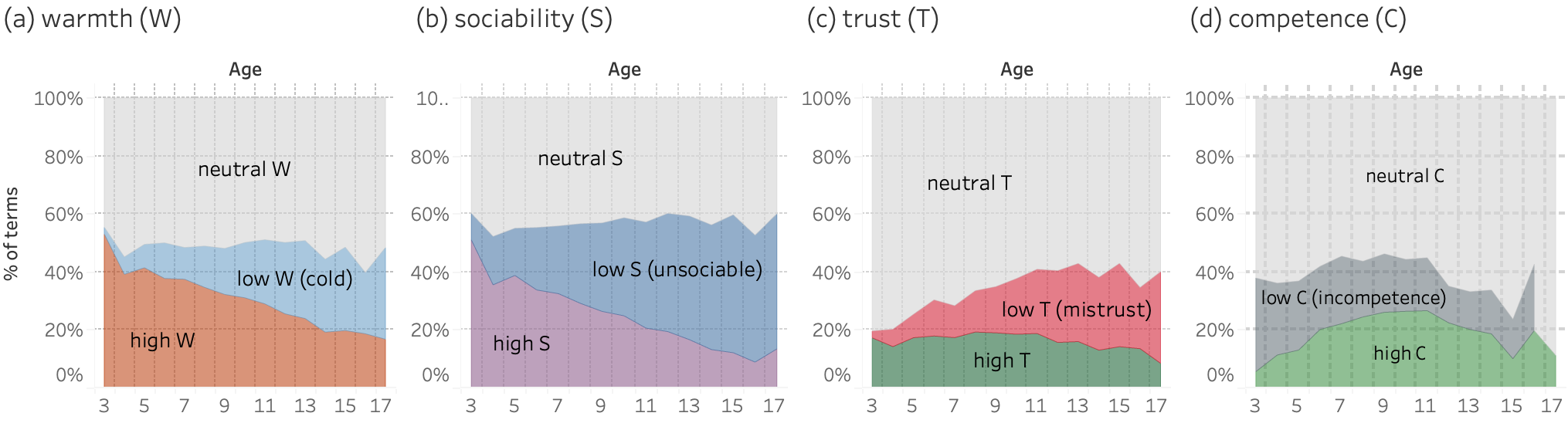}
      \caption{Stream charts of the percentages of high-, low-, and neutral WCTS words acquired in ages 3 to 17. (The three percentages for each age sum up to 100\%.)}
	  %   \caption{For each non-neutral Words of Warmth bin \textit{x} and age \textit{y}: the percentage of \textit{x} terms acquired at \textit{y} w.r.t. all non-neutral terms acquired at \textit{y}.}
	     \label{fig:wc-aoa}
      \vspace*{-3mm}
	 \end{figure*}

\section{At what rate do children acquire words associated with warmth, competence, trust, and sociability?}
\label{sec:children}
%brb% Which dimension is more important?}

% \cite{eisenbruch2022warmth}
% - warmth is prioritized over competence in multiple types of social decision-making. 
% - Existing explanations for this prioritization argue that warmth is more consequential for an observer’s welfare than is competence. We present a new explanation for the prioritization of warmth based on humans’ evolutionary history of cooperative partner choice. We argue that the prioritization of warmth evolved because ancestral humans faced greater variance in the warmth of potential cooperative partners than in their competence but greater variance in competence over time within cooperative relationships. These each made warmth more predictive than competence of the future benefits of a relationship, but because of differences in the distributions of these traits, not because of differences in their intrinsic consequentiality. 

As discussed in the Related Work, there is much we do not know about
how warmth and competence develop and which dimension is of greater significance.
To shed some light on this, we use the WCTS lexicons in combination with an age of acquisition dataset \cite{kuperman2012age} to examine when children acquire WCTS-associated words. The age of acquisition dataset includes the age at which $\sim$30K English words are commonly acquired by children.

For this set of experiments, every dimension is split into three regions: low, neutral, and high.
For WST: scores between $-$1.5 and 1.5 are considered neutral; scores $<= -1.5$ are considered low (low T or untrustworthy, low S or unsociable, low W or cold); and scores $>= 1.5$ are considered high.\footnote{For C scores (taken from the NRC VAD lexicon): %,
%brb% as suggested in \citet{mohammad-2018-obtaining}: 
scores between $-$0.33 and 0.33: neutral; between $-$1 and $-$0.33: low C; between 0.33 and 1: high C.}
% (high trust, high sociability, high warmth).   
We will refer to the set of words from the high- and low-regions as polar words. (In other words, for a given dimension, the set of polar words includes all words except the neutral words.)

Figures \ref{fig:wc-aoa} a through d show stream graphs of the percentages of high, low, and neutral words acquired each year by children.  
(For every x-axis value, the three percentages sum up to 100\%.) 
Observe that from age 3 onward, the percentage of high-W words decreases steadily with age, whereas, the percentage of low-W words increases with age.
Overall, the number of polar warmth words acquired with age stays steady at about 50\%.
The pattern for competence (d) is markedly different. The rate at which high-C words are acquired increases gradually, peaking at about 10 years of age, and then decreasing again. The rate of acquisition of low-C words is highest in the early years and decreases steadily with age. The rate of acquisition of polar words has a slight inverted U pattern (peaking at 10 years). Overall, the percent of polar W words acquired at each of the ages is higher than the percent of polar C words.

The rate of acquisition of high- and low-S words (shown in (b)) is similar to that of high- and low-W words.
The rate for high-T words starts off high and stays steady till about 10 to 11 years, after which there is a slight but steady decline. The rate of acquisition of low-T words is very small at age 3, but it increases steadily with age.

\noindent \textit{Discussion:} Overall, we see clear trends in the acquisitions of words for each of the dimensions. The  higher percentages for polar (non-neutral) warmth words vs.\@ polar competence words is consistent with the primacy of valence hypothesis (as opposed to the primacy of competence hypothesis). The markedly higher percentages for polar sociability words as opposed to polar trust words in early years, is consistent with the notion that the dimension of sociability is more important than the dimension of trust (and morality) in the early years.  Among the polar words, it is interesting that the early years are marked with a greater percentage of high-WST words, as well as low-C words. This is consistent with the notion that children receive more warmth earlier in life than later. The higher percentage of low-C words is consistent with the fact that children are more heavily dependent on caregivers in early years than in later years.

These findings have implications in developmental psychology and evolutionary linguistics. They are also relevant to understanding how children develop these key dimensions of social cognition and their role in shaping traits such as social competence and emotion regulation \cite{ROUSSOS2016133,wojciszke2009two}. 
% We hope that Words of Warmth will spur further research in these areas.

\section{Case Studies of W and C Stereotypes}
\label{sec:cases}

Our stereotypes about people often manifest in  language. \bl{The WCTS lexicons (with WTS scores from our newly created lexicon and competence scores from the NRC VAD Lexicon v2) can be used in combination with large amounts of text to %reliably infer such 
shed light on human stereotypes.}
 We make use of two methods which provide different windows into human stereotype towards various targets commonly studied in stereotype research \cite{morabito2024stop}: Direct Target Lookup of target terms in the WCTS lexicon (\textbf{Direct WCTS}) and Examining WCTS of terms co-occurring with the target terms in text (\textbf{Co-terms WCTS}).
For our Co-terms WCTS experiments we use a corpus of lemmatized and lower-cased American and Canadian geo-located posts on X (formerly Twitter) from 2015 to 2021 \cite{vishnubhotla-mohammad-2022-tusc,ABCDE}.\footnote{\bl{The TUSC tweets  \cite{vishnubhotla-mohammad-2022-tusc} with WCTS features is now part of the ABCDE dataset for Computational Affective Science \cite{ABCDE}.}}

% Specifically, we explore stereotypes and perceptions towards various targets (social groups, world leaders, objects, and ideas) (1 and 2); as well as, how ingroup and outgroup membership impacts warmth and competence associations (3 and 4).
% \begin{enumerate}
% \item Stereotypes and perceptions associated with different professions, world leaders, countries and other social groups.
% \item  Associations with different genders.
% \item  Canada--US ingroup and outgroup perceptions.
% \item  Associations of texts with ingroup and outgroup pronouns.
% \end{enumerate}
% \noindent These experiments show a sample of the wide variety of the analyses that lexicons enable.

\begin{figure}[t]
	     \centering
	     \includegraphics[width=0.5\textwidth]{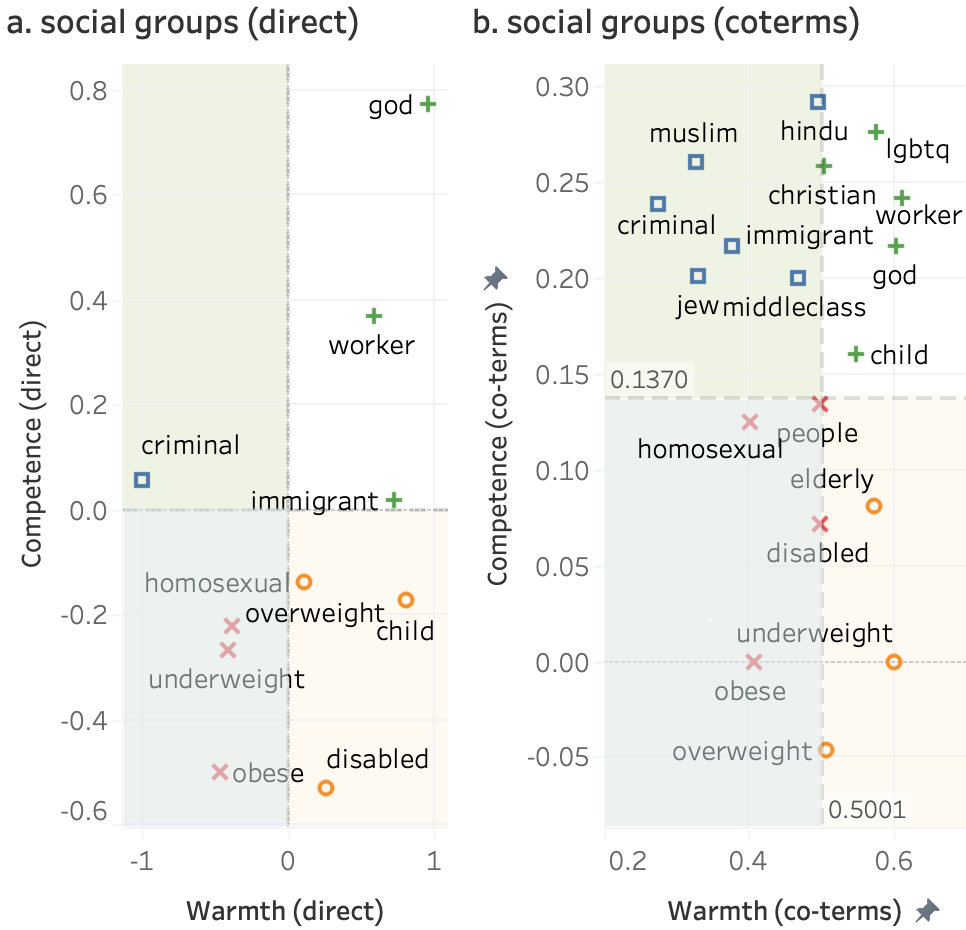}
     \vspace*{-5mm}
      \caption{W--C plots for social groups.}
	     \label{fig:groups-wc}
      %\vspace*{-1mm}
	 \end{figure}

\noindent \textbf{Direct WCTS:} One can directly look up WCTS  scores of target terms in the lexicons. For example, Figure \ref{fig:groups-wc} (a) shows the W and C scores of terms referring to various social groups. 
% The list of terms is taken from past studies on stereotype \cite{}. 
The shading marks the quadrants: 
high W and C (white), high W and low C (yellow), low W and high C (green), and low W and C (blue). Note that the average W and C scores of all 2,086 terms in the lexicons are 0.002 and 0.001, respectively (very close to 0). Therefore, since the term \textit{worker} has a positive score for both W and C, it means that people perceive \textit{worker} as being more warm and more competent than the average term in the lexicon. Note how the concept of \textit{god} is perceived as highly warm and highly competent;
\textit{disabled} as very low C; \textit{criminal} as very low W; and people outside of the socially preferred weight class as low W and C.
Some of the terms in our original list such as \textit{lgbtq, muslim,} and \textit{jew} do not occur in the lexicon. These words can be analyzed through the WCTS of their co-occurring terms described ahead.

\noindent \textbf{Co-terms WCTS:} We obtain co-occurrence based WCTS scores by examining the lexicon entries of terms co-occurring with the target terms. Steps:
\begin{compactenum}
\item Manually identify minimally ambiguous terms commonly used to refer to the target. Collect posts that include mentions of the target term(s).
E.g., using the terms \textit{nurse} and \textit{nurses} to collect posts about nursing professionals.
% Identify posts that mention the target. (We will refer to this as the \textit{target corpus}.)\\ 
% This involves: 
\item Calculate co-term  WCTS scores for each target term.
Following \cite{teodorescu-mohammad-2023-evaluating,turney-2002-thumbs} we calculate the percentage of high-W words in the target corpus minus the percentage of low-W words in the target corpus.\footnote{\citet{teodorescu-mohammad-2023-evaluating} and \citet{turney-2002-thumbs} show that this formula accurately captures the degree of emotions in various corpora for valence, arousal, anger, sadness, etc. Other similar formulae may also be be used. Our goal was to use a simple and interpretable approach that has been shown to work well for aggregate-level analysis.} CTS scores are determined similarly.
% \item Plotting relevant figures where the targets lies on the warmth--competence, trust--competence, and sociability--competence spaces.
\end{compactenum}

\begin{figure}[t]
	     \centering
	     \includegraphics[width=0.45\textwidth]{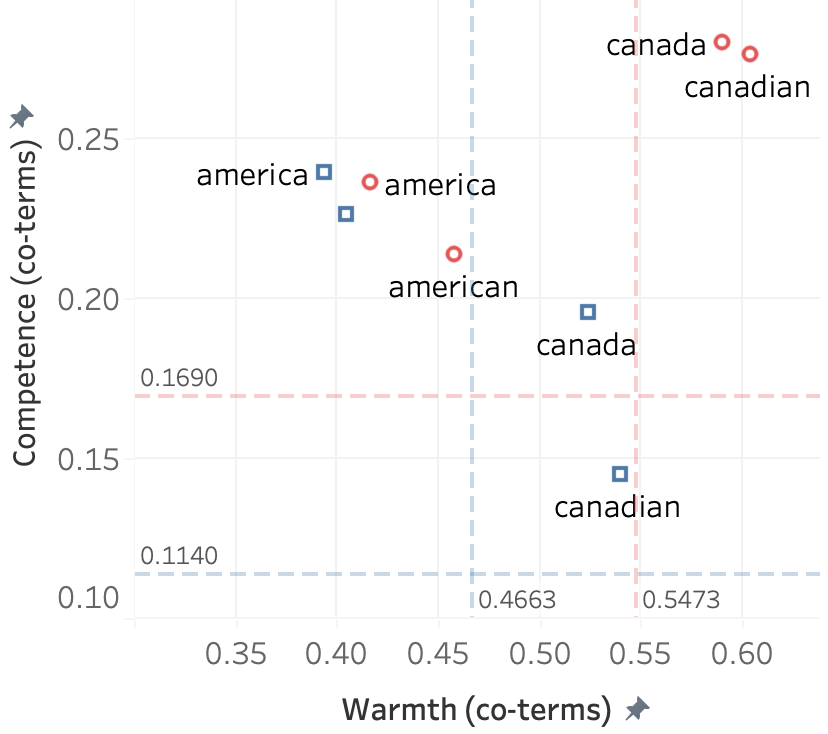}
      \caption{Co-terms W--C plot for tweets by Canadians (red) and Americans (blue) mentioning each other.}
	  \vspace*{-1mm}
         \label{fig:us-ca-wc}
      \vspace*{-1mm}
	 \end{figure}

The WCTS scores obtained using co-terms give an indication of the extent to which we use high- and low-WCTS terms in utterances that include the target term. Higher scores indicate more high-WCTS words and fewer low-WCTS words. Some important points should be noted regarding how to further interpret these scores:
\begin{compactenum}
    \item The co-term scores need not correlate with the target scores. This can happen for a number of reasons, including: how we respond when directly asked about a target may differ from our true feelings; mentions of the target may be in a restricted context not representing the full set of contexts in which the target is talked about; etc.
    \item Different groups of people may use different terms to refer to the same target entity. For example, people who use the term \textit{lgbtq} tend to view the group more positively than people who do not (e.g., those who use identity terms dis-preferred by the group).
    \item Even though the co-term scales have the same range as the direct scores ($-$1 to 1), the two metrics are not directly comparable. For one, target scores have a normal distribution around 0, whereas the co-term scores have normal distributions at an offset. For example, the average W and C scores of all 3.1 million tweets in our tweets corpus are 0.5001 and 0.1370, respectively.\footnote{This is because in a sentence we often use many warmth words, whereas even one or two coldness words are sufficient to convey a strong overall coldness tone.}
    Thus, in the analyses below we examine relative positions of targets w.r.t. the average on the W--C plots. We also consider quadrants in the W--C space with respect to the W and C averages (and not w.r.t. 0, 0). 
    % Thus, for example, we note that \textit{underweight} is in the warm and incompetent quadrant as per co-terms but in the cold and incompetent quadrant as per direct target scores.
    \item We checked for stability of the WCTS scores for a given target entity, by looking at how much the scores vary for each of the years from 2015 to 2021, and also by examining scores for morphological variants of the target term. The closeness of these scores indicate stability of the WCTS scores. For example, in Fig \ref{fig:pronouns-wc} in the Appendix we plot the values for the pronouns for every year; and in Fig \ref{fig:us-ca-wc} we plot the values for both \textit{america} and \textit{american}, as well as \textit{canada} and \textit{canadian}.  Due to limited space and for clarity we omit other plots showing scores for different years and morphological variants.
    \item We use the W--C plots as the primary mechanism to showcase the kind of analyses the WCTS lexicons enable, but note that similar analysis can be done with T--C, S--C, etc. For example, one notable aspect we found in our analysis was that while many of the social groups considered had T and S scores that were close to each other, there exist terms such as \textit{homosexual} whose T score was quite different (in this case, much lower) than their S score. This helps us understand and track how the discourse about gay people is still polarizing and a section of society uses low-trust (morality) terms when talking about this group (mirroring the known negative and harmful stereotypes against them).  
\end{compactenum}
\noindent We show some case studies below. (The Appendix has a supplemental case study of professions.)

\noindent \textbf{1. Social Groups.} Figure \ref{fig:groups-wc} (b) shows the co-terms based W and C scores of various social groups. Some notable observations include:
\begin{compactitem}
    \item \textit{muslim, jew,} and \textit{immigrant} get low-W scores (consistent with known negative stereotypes towards them in US and Canada).
    \item \textit{elderly} and \textit{underweight} get low-C and high-W scores; whereas \textit{overweight} gets an even lower C score. \textit{obese} gets low-W and low-C scores.
        \item \textit{god} gets high direct W and C scores (Figure \ref{fig:groups-wc} (a)), but the discourse around \textit{god} on X is such that the term gets lower co-terms-based C score than many other terms (b).
\end{compactitem}

% target-wc.twb

% target-wc.twb
     \begin{figure}[t]
	     \centering
	     \includegraphics[width=0.5\textwidth]{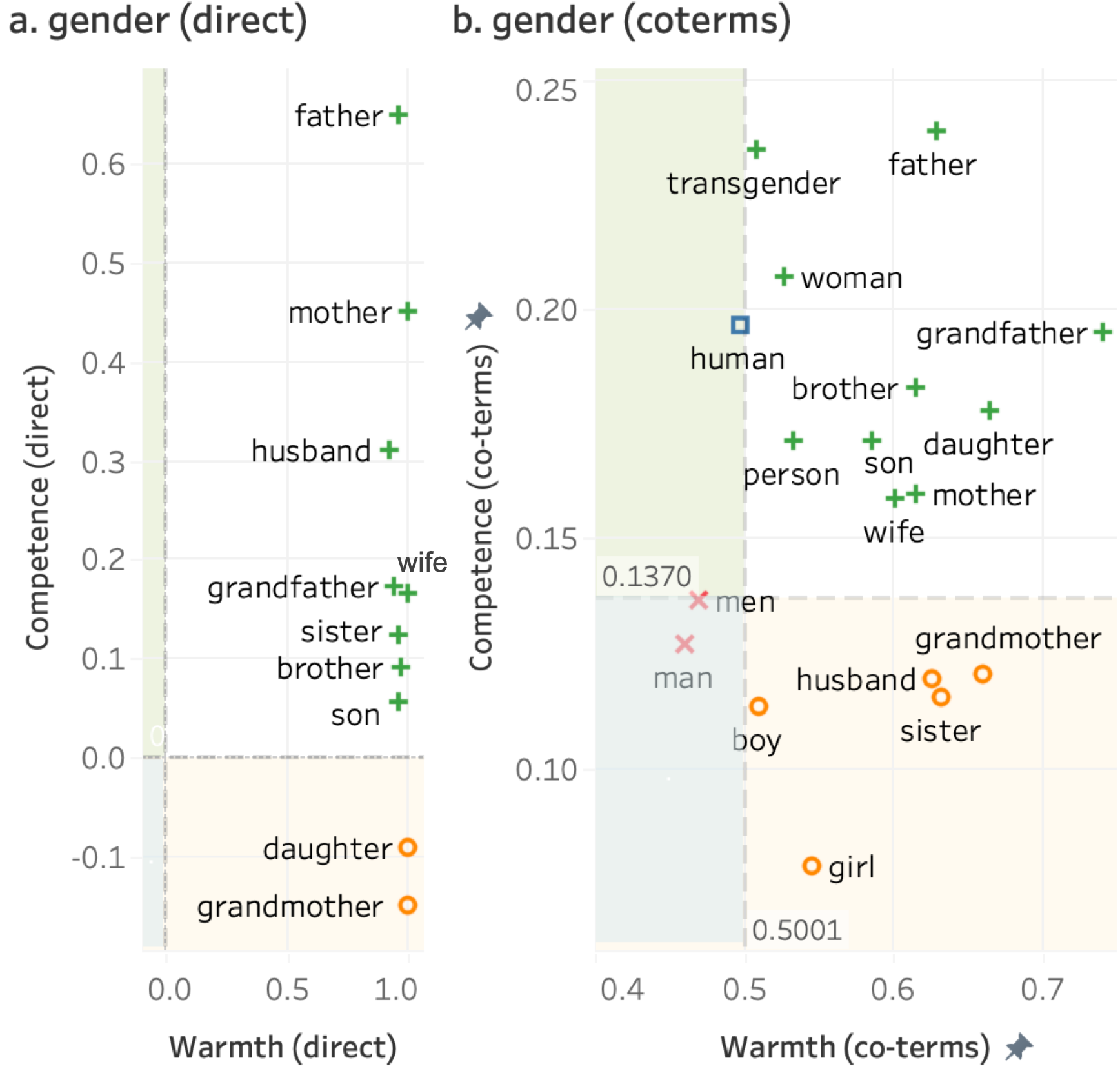}
      \vspace*{-5mm}
      \caption{W--C plots for various gender terms.}
	     \label{fig:gender-wc}
      \vspace*{-4mm}
	 \end{figure}

\noindent \textbf{2. Genders.} Figure \ref{fig:gender-wc} (a) and (b) show the direct and co-terms based W and C scores of various gender groups. Observe that:
\begin{compactitem}
    \item When asked for W and C assessment directly, people consider all these terms as high W, but perceive substantial variations in their competence.  \textit{father} and \textit{mother} are seen as high C whereas \textit{grandmother} is seen as low C.
    \item In contrast, the co-term plots show that our language has marked differences for these terms not just for C but also for W. We use the most high W and fewer low W words when \bl{mentioning \textit{grandfather}} (even more than \textit{daughter}, and \textit{grandmother}) compared to \textit{son, wife,} and \textit{husband}.  The co-term plot also shows certain additional related terms for comparison such as \textit{person} and \textit{human}. 
    \item Importantly, we see clear gender stereotypes reflected in these scores with males getting higher C scores and females getting higher W scores (with the exception of \textit{grandfather}).
\end{compactitem}

\noindent \textbf{3. Ingroup--Outgroup Impact.} Since we know which tweets were posted in Canada and which in the USA, we can use that information to examine ingroup and outgroup behavior. Specifically, we examine the W and C words used by Canadians and Americans when they refer to each other. Figure \ref{fig:us-ca-wc} shows the co-terms based W and C scores for: \textit{america, american, canada,} and \textit{canadian} obtained from the tweets by each group (Americans and Canadians). Note the blue dashed lines that indicate the average W and C scores of all posts by Americans and red dashed lines that indicate the averages for Canadians. We observe that posts by Canadians in general have higher W and C scores than posts by Americans. 
Further, the scores provide some evidence that Canadians view themselves as more competent and much warmer than their neighbours 
% down south 
(consistent with ingroup and outgroup stereotypes).  In contrast, while Americans view themselves as more competent than Canadians, they too perceive Canadians as warmer (suggesting that the \textit{Canadians are nicer} stereotype overrides the outgroup stereotype in this case). 

% target-wc.twb

\noindent \textbf{4. Pronouns.} Analyzing posts with pronouns gives us interesting insights into how we speak about ourselves (1st person) vs.\@ a person we are directly engaging with (2nd person) vs.\@ a third person. How do the W and C scores differ for these different types of posts? Figure \ref{fig:pronouns-wc} in the Appendix 
shows the plots. % and we list the key findings below: 
\begin{compactitem}
    \item Firstly, we note that different pronouns fall in markedly separate areas of the W--C plots (can also be seen from the yearly plot shown in (b)).
    \item Secondly, first-person-singular mentions (\textit{I} and \textit{me}) are associated with low C, whereas and second person pronoun mentions (\textit{you}) are associated with high C.
   % \cite{kacewicz2014pronoun,pennebaker2011secret}.
This is consistent with findings by \citet{kacewicz2014pronoun} and \cite{pennebaker2011secret}, who show that \textit{I} is often overused by people in low-power/status relations, whereas \textit{you} and \textit{we} are often used by people in high-power/status relationships.
Additionally, we speculate this is because social media interactions often involve people who may not personally know each other, and so as a matter of being socially congenial to the addressee we prioritize competence over warmth and tend to use more high-C and fewer low-C words.
    \item Mentions of \textit{we} are associated with high W.  This is consistent with findings that show that \textit{we} commonly occurs in more positive contexts  \cite{senden2014biases,pennebaker2011secret}.
\end{compactitem}
  %    \begin{figure*}[t]
	 %     \centering
	 %     \includegraphics[width=\textwidth]{figures/pronouns-wc.png}
  %     \caption{Direct and co-term W--C plots for various pronouns.}
	 %     \label{fig:pronouns-wc}
  %     \vspace*{-3mm}
	 % \end{figure*}
% \subsection{Warmth and Competence Associations of Genders}
% \subsection{Canada--US Ingroup and Outgroup Warmth and Competence Perceptions}
% \subsection{Warmth and Competence Associations of Ingroup and Outgroup Pronouns}
% \noindent 
\noindent With the lexicons made freely available, we hope they will spur further detailed exploration into the targets discussed above as well as numerous others.

\section{Conclusion}
We created Words of Warmth --- a large lexicon of word--warmth, word--trust, and word--sociability association scores from over 700,000 responses from hundreds of respondents.
We showed that the scores are highly reliable (>0.94 SHR).  We used the lexicon %along with an existing competence lexicon 
to study the rate at which WCTS words are acquired with age. 
Finally, we presented case studies on how the lexicons can be used to track stereotypes. % in language. 
% We make Words of Warmth freely available to enable a wide variety of anxiety associated research. 
% Finally, we presented case studies of warmth and competence perceptions reflected in US and Canadian social media posts towards targets commonly studied in stereotype research.

We make the lexicons freely available to foster further research, notably on understanding: how human beings develop their inner stereotype model; how early childhood experiences can impact our stereotype model, social competence, emotion regulation, and personality traits; and tracing stereotype and bias of a source population of interest towards a target entity of interest.
%We also believe future work can make extensive use of the lexicon in detailed stereotype and bias analysis of a source population of interest towards a target entity of interest. 
% Some areas of particular interest
Other areas of future work, include: using the lexicons to study text produced by generative AI (the degree of stereotypes it reflects and under what conditions); 
% exploring when and why trust and sociability towards a target of interest diverge ; 
creating warmth lexicons for more languages and cultures to enable cross-cultural comparisons; developing automatic systems to assess systematic and consistent trends in WCTS biases on various social media channels (such as various subreddits); and analyzing one's own writing (over a period of time) to understand how we assess ourselves in terms of WCTS (and its implications on mental health).
One can even use the lexicons to assess perceptions of complex social issues such as how we should deal with climate change and immigration.
Thus, we see wide applicability of Words of Warmth in psychology, computational affective science, NLP, public health, digital humanities, political science, and social science research. 

% (such as in the seven bullets listed in the introduction) in NLP, psychology, social science, etc.

% most notably in understanding and managing anxiety associated disorders, 

\section{Limitations}
\label{sec:limitations}

The large body of social psychology work on the dimensions of social cognition and stereotype are based on human responses. Traditionally, these studies only included responses from people in the western world. This can lead to over generalizations. However, more recently there has been growing work that confirms the importance of warmth and competence across cultures \cite{fiske2016stereotype,grigoryev2019mapping} and showing the evolutionary basis of these dimensions \cite{macdonald1992warmth,cuddy2007bias,eisenbruch2022warmth}. Nonetheless, it is entirely possible, that in some cultures W and C are not the primary dimensions of social cognition. 

This work develops WTS lexicons for English, based on responses primarily from the US, Canada, UK, and India. Thus it is important to contextualize any conclusions as those applying to English speakers, and that too mainly US speakers. Just as the social psychology work, true global conclusions can only be drawn from many such works on many languages and cultures. We see this work as a first step that paves the way for more work in various other languages and cultures. 
% Further, the lexicon will be useful in the analysis of North American stereotypes, even if it needs some adjustment when applied to text from elsewhere.

\noindent See discussions of limitations in how the lexicons can be used and interpreted in the Ethics Statement below (\S \ref{sec:ethics}).

\section{Ethics and Data Statement}
\label{sec:ethics}

The crowd-sourced task presented in this paper was approved by our Institutional Research Ethics Board. 
Our annotation process stored no information about annotator identity and as such there is no privacy risk to them.  The individual words selected did not pose any risks beyond the risks of occasionally reading text on the internet. 
The annotators were free to do as many word annotations as they wished. The instructions included a brief description of the purpose of the task (Figures \ref{fig:trust-det-instr} and \ref{fig:soc-det-instr}).

WCTS assessments are complex, nuanced, and often instantaneous mental judgments. Additionally, each individual may use language to convey these assessments slightly differently.
% See \citet{Mohammad23ethicslex} for a discussion of good practices and ethical considerations when using emotion lexicons. See \citet{Mohammad22AER} for a broader discussion of ethical considerations relevant to automatic emotion recognition.
We discuss below notable ethical considerations when computationally analyzing WCTS through language. 

Importantly, Words of Warmth should not be used as a standalone tool for detecting stereotypes and bias in individual utterances. At minimum, it must be used in combination with various other sources of information, large amounts of texts, and appropriate contextualization (the same text may mean different things in different contexts). See considerations below, which also apply broadly to any lexical dataset of association norms (many of these build on similar issues for emotions, discussed in \citet{Mohammad23ethicslex,Mohammad22AER}): 
% BB We list some notable ones below. 
% Many of these were first introduced in \cite{mohammad2020practical,mohammad2023LexEthics}. We adapted them to anxiety association and added to the discussion.

\begin{compactenum}
    \item \textit{Coverage:} We sampled a large number of English words from other lexical sources (which themselves sample from many sources). Yet, the words included do not cover all domains, genres, and people of different locations, socio-economic strata, etc.\@ equally. It likely includes more of the vocabulary common in the United States with socio-economic and educational backgrounds that allow for technology access.
 
    \item \textit{Word Senses and Dominant Sense Priors:} Words when used in different senses and contexts may be associated with different degrees of WCTS associations. The entries in Words of Warmth are indicative of the associations with the predominant senses of the words. This is usually not problematic because most words have a highly dominant main sense (which occurs much more frequently than the other senses). 
In specialized domains, some terms might have a different dominant sense than in general usage. Entries in the lexicon for such terms should be appropriately updated or removed. 
Further, any conclusions using the lexicon should be made based on relative change of associations using a large number of textual tokens. For example, if there is a marked increase in coldness words from one period to the next, where each period has thousands of word tokens, then the impact of word sense ambiguity is minimal, and it is likely that some broader phenomenon is causing the marked increase in coldness words. (See last two bullets.)

    \item \textit{Not Immutable:} The WCTS scores do not indicate an inherent unchangeable attribute. The associations can change with time (e.g., the decrease in coldness and immorality associated with \textit{inter-race relationships} over the last 100 years), but the lexicon entries are largely fixed. They pertain to the time they are created. However, they can be updated with time.
  
    \item \textit{Socio-Cultural Biases:} The annotations for WCTS capture various human biases. These biases may be systematically different for different socio-cultural groups. Our data was annotated by mostly US, Canadian, UK, and Indian English speakers, but even within these countries there are many diverse socio-cultural groups.
    Notably, crowd annotators on Amazon Mechanical Turk do not reflect populations at large. In the US for example, they tend to skew towards male, white, and younger people. However, compared to studies that involve just a handful of annotators, crowd annotations benefit from drawing on hundreds and thousands of annotators (such as this work). 
 
    \item \textit{Inappropriate Biases:} Our biases impact how we view the world, and some of the biases of an individual may be inappropriate. For example, one may have race or gender-related biases that may percolate subtly into one's notions of WCTS associated with words. 
    Our dataset curation was careful to avoid words from problematic sources. We also ask people annotate terms based on what most English speakers think (as opposed to what they themselves think). This helps to some extent, but the lexicon may still capture some historical WTS associations with certain identity groups. This can  be useful for some socio-cultural studies; but we also caution that WCTS associations with identity groups be carefully contextualized to avoid false conclusions.    
  
    \item \textit{Perceptions (not “right” or “correct” labels):} Our goal here was to identify common perceptions of WTS association. These are not meant to be ``correct'' or ``right'' answers, but rather what the majority of the annotators believe based on their intuitions of the English language.
   
    \item \textit{Avoid Essentialism:} When using the lexicon alone, it is more appropriate to make claims about WCTS word usage rather than the WCTS of the speakers. For example, {\it `the use of trust words in the context of the target group grew by 20\%'} rather than {\it `trust in the target group grew by 20\%'}. In certain contexts, and with additional information, the inferences from word usage can be used to make broader claims. 
\item \textit{Avoid Overclaiming:} Inferences drawn from larger amounts of text are often more reliable than those drawn from small amounts of text.
 For example, {\it `the use of warmth words grew by 20\%'} is informative when determined from hundreds, thousands, tens of thousands, or more instances. Do not draw inferences about a single sentence or utterance from the WCTS associations of its constituent words.
\item \textit{Embrace Comparative Analyses:} Comparative analyses can be much more useful than stand-alone analyses. Often, WCTS word counts and percentages on their own are not very useful. 
For example, {\it `the use of warmth words grew by 20\% when compared to [data from last year, data from a different person, etc.]'} is more useful than saying {\it `on average, 5 warmth words were used in every 100 words'}.
\end{compactenum}
\noindent We recommend careful reflection of ethical considerations relevant for the specific context of deployment when using Words of Warmth.

\section*{Acknowledgments}
\bl{Thanks to Susan Fiske, Tara Small, Nedjma Oousidhoum, Jan Philip Wahle, and Kathleen Fraser for helpful discussions. Thanks to Jan Philip Wahle for the early access to the ABCDE dataset.}
\vspace*{-4mm}

\bibliography{anthology,custom}

\begin{thebibliography}{54}
\expandafter\ifx\csname natexlab\endcsname\relax\def\natexlab#1{#1}\fi

\bibitem[{Abele et~al.(2016)Abele, Hauke, Peters, Louvet, Szymkow, and Duan}]{abele2016facets}
Andrea~E Abele, Nicole Hauke, Kim Peters, Eva Louvet, Aleksandra Szymkow, and Yanping Duan. 2016.
\newblock Facets of the fundamental content dimensions: Agency with competence and assertiveness—communion with warmth and morality.
\newblock \emph{Frontiers in psychology}, 7:1810.

\bibitem[{Altschul et~al.(2016)Altschul, Lee, and Gershoff}]{altschul2016hugs}
Inna Altschul, Shawna~J Lee, and Elizabeth~T Gershoff. 2016.
\newblock Hugs, not hits: Warmth and spanking as predictors of child social competence.
\newblock \emph{Journal of Marriage and Family}, 78(3):695--714.

\bibitem[{Ariza-Casabona et~al.(2022)Ariza-Casabona, Schmeisser-Nieto, Nofre, Taul{\'e}, Amig{\'o}, Chulvi, and Rosso}]{ariza2022overview}
Alejandro Ariza-Casabona, Wolfgang~S Schmeisser-Nieto, Montserrat Nofre, Mariona Taul{\'e}, Enrique Amig{\'o}, Berta Chulvi, and Paolo Rosso. 2022.
\newblock Overview of detests at iberlef 2022: Detection and classification of racial stereotypes in spanish.
\newblock \emph{Procesamiento del lenguaje natural}, 69:217--228.

\bibitem[{Baines et~al.(2024)Baines, Gruia, Collyer-Hoar, and Rubegni}]{baines2024playgrounds}
Alexander Baines, Lidia Gruia, Gail Collyer-Hoar, and Elisa Rubegni. 2024.
\newblock Playgrounds and prejudices: Exploring biases in generative ai for children.
\newblock In \emph{Proceedings of the 23rd Annual ACM Interaction Design and Children Conference}, pages 839--843.

\bibitem[{Blodgett et~al.(2020)Blodgett, Barocas, Daum{\'e}~III, and Wallach}]{blodgett2020language}
Su~Lin Blodgett, Solon Barocas, Hal Daum{\'e}~III, and Hanna Wallach. 2020.
\newblock Language (technology) is power: A critical survey of" bias" in nlp.
\newblock \emph{arXiv preprint arXiv:2005.14050}.

\bibitem[{Bodenhausen et~al.(2012)Bodenhausen, Kang, and Peery}]{bodenhausen2012social}
Galen~V Bodenhausen, Sonia~K Kang, and Destiny Peery. 2012.
\newblock Social categorization and the perception of social groups.
\newblock \emph{The Sage handbook of social cognition}, pages 311--329.

\bibitem[{Bosco et~al.(2023)Bosco, Patti, Frenda, Cignarella, Paciello, and D’Errico}]{bosco2023detecting}
Cristina Bosco, Viviana Patti, Simona Frenda, Alessandra~Teresa Cignarella, Marinella Paciello, and Francesca D’Errico. 2023.
\newblock Detecting racial stereotypes: An italian social media corpus where psychology meets nlp.
\newblock \emph{Information Processing \& Management}, 60(1):103118.

\bibitem[{Caliskan et~al.(2017)Caliskan, Bryson, and Narayanan}]{caliskan2017semantics}
Aylin Caliskan, Joanna~J Bryson, and Arvind Narayanan. 2017.
\newblock Semantics derived automatically from language corpora contain human-like biases.
\newblock \emph{Science}, 356(6334):183--186.

\bibitem[{Cuddy et~al.(2007)Cuddy, Fiske, and Glick}]{cuddy2007bias}
Amy~JC Cuddy, Susan~T Fiske, and Peter Glick. 2007.
\newblock The bias map: behaviors from intergroup affect and stereotypes.
\newblock \emph{Journal of personality and social psychology}, 92(4):631.

\bibitem[{Cuddy et~al.(2011)Cuddy, Glick, and Beninger}]{cuddy2011dynamics}
Amy~JC Cuddy, Peter Glick, and Anna Beninger. 2011.
\newblock The dynamics of warmth and competence judgments, and their outcomes in organizations.
\newblock \emph{Research in organizational behavior}, 31:73--98.

\bibitem[{Durante and Fiske(2017)}]{durante2017social}
Federica Durante and Susan~T Fiske. 2017.
\newblock How social-class stereotypes maintain inequality.
\newblock \emph{Current opinion in psychology}, 18:43--48.

\bibitem[{Eisenbruch and Krasnow(2022)}]{eisenbruch2022warmth}
Adar~B Eisenbruch and Max~M Krasnow. 2022.
\newblock Why warmth matters more than competence: A new evolutionary approach.
\newblock \emph{Perspectives on Psychological Science}, 17(6):1604--1623.

\bibitem[{Fiske et~al.(2002)Fiske, Cuddy, Glick, and Xu}]{fiske2002}
Susan Fiske, Amy Cuddy, Peter Glick, and Jun Xu. 2002.
\newblock \href {https://doi.org/10.1037/0022-3514.82.6.878} {A model of (often mixed) stereotype content: Competence and warmth respectively follow from perceived status and competition}.
\newblock \emph{Journal of Personality and Social Psychology}, 82:878--902.

\bibitem[{Fiske(2018)}]{fiske2018stereotype}
Susan~T Fiske. 2018.
\newblock Stereotype content: Warmth and competence endure.
\newblock \emph{Current directions in psychological science}, 27(2):67--73.

\bibitem[{Fiske and Durante(2016)}]{fiske2016stereotype}
Susan~T Fiske and Federica Durante. 2016.
\newblock Stereotype content across cultures.
\newblock \emph{Handbook of advances in culture and psychology}, 6:209--258.

\bibitem[{Fiske et~al.(2014)Fiske, Durante et~al.}]{fiske2014never}
Susan~T Fiske, Federica Durante, et~al. 2014.
\newblock Never trust a politician? collective distrust, relational accountability, and voter response.
\newblock \emph{Power, politics, and paranoia: Why people are suspicious of their leaders}, pages 91--105.

\bibitem[{Fraser et~al.(2024)Fraser, Kiritchenko, and Nejadgholi}]{fraser-etal-2024-stereotype}
Kathleen Fraser, Svetlana Kiritchenko, and Isar Nejadgholi. 2024.
\newblock \href {https://doi.org/10.18653/v1/2024.starsem-1.2} {How does stereotype content differ across data sources?}
\newblock In \emph{Proceedings of the 13th Joint Conference on Lexical and Computational Semantics (*SEM 2024)}, pages 18--34, Mexico City, Mexico. Association for Computational Linguistics.

\bibitem[{Grigoryev et~al.(2019)Grigoryev, Fiske, and Batkhina}]{grigoryev2019mapping}
Dmitry Grigoryev, Susan~T Fiske, and Anastasia Batkhina. 2019.
\newblock Mapping ethnic stereotypes and their antecedents in russia: The stereotype content model.
\newblock \emph{Frontiers in psychology}, 10:1643.

\bibitem[{Hilton and Von~Hippel(1996)}]{hilton1996stereotypes}
James~L Hilton and William Von~Hippel. 1996.
\newblock Stereotypes.
\newblock \emph{Annual review of psychology}, 47(1):237--271.

\bibitem[{Kacewicz et~al.(2014)Kacewicz, Pennebaker, Davis, Jeon, and Graesser}]{kacewicz2014pronoun}
Ewa Kacewicz, James~W Pennebaker, Matthew Davis, Moongee Jeon, and Arthur~C Graesser. 2014.
\newblock Pronoun use reflects standings in social hierarchies.
\newblock \emph{Journal of Language and Social Psychology}, 33(2):125--143.

\bibitem[{Kiritchenko and Mohammad(2018)}]{kiritchenko-mohammad-2018-examining}
Svetlana Kiritchenko and Saif Mohammad. 2018.
\newblock \href {https://doi.org/10.18653/v1/S18-2005} {Examining gender and race bias in two hundred sentiment analysis systems}.
\newblock In \emph{Proceedings of the Seventh Joint Conference on Lexical and Computational Semantics}, pages 43--53, New Orleans, Louisiana. Association for Computational Linguistics.

\bibitem[{Koch et~al.(2024)Koch, Smith, Fiske, Abele, Ellemers, and Yzerbyt}]{koch2024validating}
Alex Koch, Austin Smith, Susan~T Fiske, Andrea~E Abele, Naomi Ellemers, and Vincent Yzerbyt. 2024.
\newblock Validating a brief measure of four facets of social evaluation.
\newblock \emph{Behavior Research Methods}, 56(8):8521--8539.

\bibitem[{Koenig and Echols(2003)}]{koenig2003infants}
Melissa~A Koenig and Catharine~H Echols. 2003.
\newblock Infants' understanding of false labeling events: The referential roles of words and the speakers who use them.
\newblock \emph{Cognition}, 87(3):179--208.

\bibitem[{Kotek et~al.(2023)Kotek, Dockum, and Sun}]{kotek2023gender}
Hadas Kotek, Rikker Dockum, and David Sun. 2023.
\newblock Gender bias and stereotypes in large language models.
\newblock In \emph{Proceedings of the ACM collective intelligence conference}, pages 12--24.

\bibitem[{Kuperman et~al.(2012)Kuperman, Stadthagen-Gonzalez, and Brysbaert}]{kuperman2012age}
Victor Kuperman, Hans Stadthagen-Gonzalez, and Marc Brysbaert. 2012.
\newblock Age-of-acquisition ratings for 30,000 english words.
\newblock \emph{Behavior research methods}, 44:978--990.

\bibitem[{MacDonald(1992)}]{macdonald1992warmth}
Kevin MacDonald. 1992.
\newblock Warmth as a developmental construct: An evolutionary analysis.
\newblock \emph{Child development}, 63(4):753--773.

\bibitem[{Maina et~al.(2018)Maina, Belton, Ginzberg, Singh, and Johnson}]{maina2018decade}
Ivy~W Maina, Tanisha~D Belton, Sara Ginzberg, Ajit Singh, and Tiffani~J Johnson. 2018.
\newblock A decade of studying implicit racial/ethnic bias in healthcare providers using the implicit association test.
\newblock \emph{Social science \& medicine}, 199:219--229.

\bibitem[{Mohammad(2018)}]{mohammad-2018-obtaining}
Saif Mohammad. 2018.
\newblock \href {https://doi.org/10.18653/v1/P18-1017} {Obtaining reliable human ratings of valence, arousal, and dominance for 20,000 {E}nglish words}.
\newblock In \emph{Proceedings of the 56th Annual Meeting of the Association for Computational Linguistics (Volume 1: Long Papers)}, pages 174--184, Melbourne, Australia. Association for Computational Linguistics.

\bibitem[{Mohammad(2022)}]{Mohammad22AER}
Saif~M. Mohammad. 2022.
\newblock Ethics sheet for automatic emotion recognition and sentiment analysis.
\newblock \emph{Computational Linguistics}, 48(2):239--278.

\bibitem[{Mohammad(2023)}]{Mohammad23ethicslex}
Saif~M. Mohammad. 2023.
\newblock Best practices in the creation and use of emotion lexicons.
\newblock In \emph{Proceedings of the 17th Conference of the European Chapter of the Association for Computational Linguistics}, Dubrovnik, Croatia. Association for Computational Linguistics.

\bibitem[{Mohammad(2024)}]{mohammad-2024-worrywords}
Saif~M. Mohammad. 2024.
\newblock \href {https://doi.org/10.18653/v1/2024.emnlp-main.910} {{W}orry{W}ords: Norms of anxiety association for over 44k {E}nglish words}.
\newblock In \emph{Proceedings of the 2024 Conference on Empirical Methods in Natural Language Processing}, pages 16261--16278, Miami, Florida, USA. Association for Computational Linguistics.

\bibitem[{Mohammad(2025)}]{vad-v2}
Saif~M. Mohammad. 2025.
\newblock \href {https://arxiv.org/abs/2503.23547} {{NRC VAD Lexicon v2: Norms for Valence, Arousal, and Dominance for over 55k English Terms}}.
\newblock \emph{arXiv preprint arXiv:2503.23547}.

\bibitem[{Moors et~al.(2013)Moors, De~Houwer, Hermans, Wanmaker, Van~Schie, Van~Harmelen, De~Schryver, De~Winne, and Brysbaert}]{moors2013norms}
Agnes Moors, Jan De~Houwer, Dirk Hermans, Sabine Wanmaker, Kevin Van~Schie, Anne-Laura Van~Harmelen, Maarten De~Schryver, Jeffrey De~Winne, and Marc Brysbaert. 2013.
\newblock Norms of valence, arousal, dominance, and age of acquisition for 4,300 dutch words.
\newblock \emph{Behavior research methods}, 45(1):169--177.

\bibitem[{Morabito et~al.(2024)Morabito, Madhusudan, McDonald, and Emami}]{morabito2024stop}
Robert Morabito, Sangmitra Madhusudan, Tyler McDonald, and Ali Emami. 2024.
\newblock Stop! benchmarking large language models with sensitivity testing on offensive progressions.
\newblock \emph{arXiv preprint arXiv:2409.13843}.

\bibitem[{Nicolas et~al.(2021)Nicolas, Bai, and Fiske}]{nicolas2021comprehensive}
Gandalf Nicolas, Xuechunzi Bai, and Susan~T Fiske. 2021.
\newblock Comprehensive stereotype content dictionaries using a semi-automated method.
\newblock \emph{European Journal of Social Psychology}, 51(1):178--196.

\bibitem[{Nosek et~al.(2005)Nosek, Greenwald, and Banaji}]{nosek2005understanding}
Brian~A Nosek, Anthony~G Greenwald, and Mahzarin~R Banaji. 2005.
\newblock Understanding and using the implicit association test: Ii. method variables and construct validity.
\newblock \emph{Personality and Social Psychology Bulletin}, 31(2):166--180.

\bibitem[{Pennebaker(2011)}]{pennebaker2011secret}
James~W Pennebaker. 2011.
\newblock The secret life of pronouns.
\newblock \emph{New Scientist}, 211(2828):42--45.

\bibitem[{Roussos and Dunham(2016)}]{ROUSSOS2016133}
Gina Roussos and Yarrow Dunham. 2016.
\newblock \href {https://doi.org/https://doi.org/10.1016/j.jecp.2015.08.009} {The development of stereotype content: The use of warmth and competence in assessing social groups}.
\newblock \emph{Journal of Experimental Child Psychology}, 141:133--144.

\bibitem[{S{\'a}nchez-Junquera et~al.(2021)S{\'a}nchez-Junquera, Chulvi, Rosso, and Ponzetto}]{sanchez2021you}
Javier S{\'a}nchez-Junquera, Berta Chulvi, Paolo Rosso, and Simone~Paolo Ponzetto. 2021.
\newblock How do you speak about immigrants? taxonomy and stereoimmigrants dataset for identifying stereotypes about immigrants.
\newblock \emph{Applied Sciences}, 11(8):3610.

\bibitem[{Schmeisser-Nieto et~al.(2024)Schmeisser-Nieto, Cignarella, Bourgeade, Frenda, Ariza-Casabona, Laurent, Cicirelli, Marra, Corbelli, Benamara et~al.}]{schmeisser2024stereohoax}
Wolfgang~S Schmeisser-Nieto, Alessandra~Teresa Cignarella, Tom Bourgeade, Simona Frenda, Alejandro Ariza-Casabona, Mario Laurent, Paolo~Giovanni Cicirelli, Andrea Marra, Giuseppe Corbelli, Farah Benamara, et~al. 2024.
\newblock Stereohoax: a multilingual corpus of racial hoaxes and social media reactions annotated for stereotypes.
\newblock \emph{Language Resources and Evaluation}, pages 1--39.

\bibitem[{Send{\'e}n et~al.(2014)Send{\'e}n, Lindholm, and Sikstr{\"o}m}]{senden2014biases}
Marie~Gustafsson Send{\'e}n, Torun Lindholm, and Sverker Sikstr{\"o}m. 2014.
\newblock Biases in news media as reflected by personal pronouns in evaluative contexts.
\newblock \emph{Social Psychology}.

\bibitem[{Swencionis et~al.(2017)Swencionis, Dupree, and Fiske}]{swencionis2017warmth}
Jillian~K Swencionis, Cydney~H Dupree, and Susan~T Fiske. 2017.
\newblock Warmth-competence tradeoffs in impression management across race and social-class divides.
\newblock \emph{Journal of Social Issues}, 73(1):175--191.

\bibitem[{Tan and Celis(2019)}]{tan2019assessing}
Yi~Chern Tan and L~Elisa Celis. 2019.
\newblock Assessing social and intersectional biases in contextualized word representations.
\newblock \emph{Advances in neural information processing systems}, 32.

\bibitem[{Teodorescu and Mohammad(2023)}]{teodorescu-mohammad-2023-evaluating}
Daniela Teodorescu and Saif Mohammad. 2023.
\newblock \href {https://doi.org/10.18653/v1/2023.findings-emnlp.271} {Evaluating emotion arcs across languages: Bridging the global divide in sentiment analysis}.
\newblock In \emph{Findings of the Association for Computational Linguistics: EMNLP 2023}, pages 4124--4137, Singapore. Association for Computational Linguistics.

\bibitem[{Thelwall(2018)}]{thelwall2018gender}
Mike Thelwall. 2018.
\newblock Gender bias in sentiment analysis.
\newblock \emph{Online Information Review}, 42(1):45--57.

\bibitem[{Tummeltshammer et~al.(2014)Tummeltshammer, Wu, Sobel, and Kirkham}]{tummeltshammer2014infants}
Kristen~Swan Tummeltshammer, Rachel Wu, David~M Sobel, and Natasha~Z Kirkham. 2014.
\newblock Infants track the reliability of potential informants.
\newblock \emph{Psychological science}, 25(9):1730--1738.

\bibitem[{Turney(2002)}]{turney-2002-thumbs}
Peter Turney. 2002.
\newblock \href {https://doi.org/10.3115/1073083.1073153} {Thumbs up or thumbs down? semantic orientation applied to unsupervised classification of reviews}.
\newblock In \emph{Proceedings of the 40th Annual Meeting of the Association for Computational Linguistics}, pages 417--424, Philadelphia, Pennsylvania, USA. Association for Computational Linguistics.

\bibitem[{Vishnubhotla and Mohammad(2022)}]{vishnubhotla-mohammad-2022-tusc}
Krishnapriya Vishnubhotla and Saif~M. Mohammad. 2022.
\newblock \href {https://aclanthology.org/2022.lrec-1.442/} {{Tweet Emotion Dynamics}: Emotion word usage in tweets from {US} and {C}anada}.
\newblock In \emph{Proceedings of the Thirteenth Language Resources and Evaluation Conference}, pages 4162--4176, Marseille, France. European Language Resources Association.

\bibitem[{V{\~o} et~al.(2009)V{\~o}, Conrad, Kuchinke, Urton, Hofmann, and Jacobs}]{vo2009berlin}
Melissa~LH V{\~o}, Markus Conrad, Lars Kuchinke, Karolina Urton, Markus~J Hofmann, and Arthur~M Jacobs. 2009.
\newblock The berlin affective word list reloaded (bawl-r).
\newblock \emph{Behavior research methods}, 41(2):534--538.

\bibitem[{Wahle et~al.(2025)Wahle, Vishnubhotla, Gipp, and Mohammad}]{ABCDE}
Jan~Philip Wahle, Krishnapriya Vishnubhotla, Bela Gipp, and Saif~M. Mohammad. 2025.
\newblock Affect, body, cognition, demographics, and emotion: The abcde of text features for computational affective science.
\newblock \emph{arXiv}.

\bibitem[{Warriner et~al.(2013)Warriner, Kuperman, and Brysbaert}]{warriner2013norms}
Amy~Beth Warriner, Victor Kuperman, and Marc Brysbaert. 2013.
\newblock Norms of valence, arousal, and dominance for 13,915 {E}nglish lemmas.
\newblock \emph{Behavior Research Methods}, 45(4):1191--1207.

\bibitem[{Weir(2005)}]{weir2005quantifying}
Joseph~P Weir. 2005.
\newblock Quantifying test-retest reliability using the intraclass correlation coefficient and the sem.
\newblock \emph{The Journal of Strength \& Conditioning Research}, 19(1):231--240.

\bibitem[{Wojciszke et~al.(2009)Wojciszke, Abele, and Baryla}]{wojciszke2009two}
Bogdan Wojciszke, Andrea~E Abele, and Wies{\l}aw Baryla. 2009.
\newblock Two dimensions of interpersonal attitudes: Liking depends on communion, respect depends on agency.
\newblock \emph{European Journal of Social Psychology}, 39(6):973--990.

\bibitem[{Zhou et~al.(2024)Zhou, Abhishek, Derdenger, Kim, and Srinivasan}]{zhou2024bias}
Mi~Zhou, Vibhanshu Abhishek, Timothy Derdenger, Jaymo Kim, and Kannan Srinivasan. 2024.
\newblock Bias in {G}enerative {AI}.
\newblock \emph{arXiv preprint arXiv:2403.02726}.

\end{thebibliography}
\bibliographystyle{acl_natbib}

\newpage
\appendix

\section{APPENDIX}
\label{appendix-a}

\begin{table}[t]
\centering
% \begin{center}
{\small
\begin{tabular}{lrrr}
\hline
\bf Word	&\bf Sociability &\bf Trust &\bf Warmth \\\hline
consoler	&3.00	&2.00	&3.00\\
cohesiveness	&3.00	&2.18	&3.00\\
wedding	&2.88	&2.22	&2.88\\
blessed	&2.83	&2.27	&2.83\\
conversant	&2.75	&0.89	&2.75\\
folk	&2.67	&1.30	&2.67\\
luckiest	&2.57	&0.27	&2.57\\
ethicist	&1.00	&2.50	&2.50\\
epidemiologist	&-0.71	&2.36	&2.36\\
neuropsychologist	&-0.62	&2.27	&2.27\\
sumptuously	&2.14	&0.55	&2.14\\
dauntless	&2.00	&1.73	&2.00\\
grief	&2.00	&0.20	&2.00\\
sundeck	&1.88	&0.27	&1.88\\
schoolbook	&1.14	&1.80	&1.80\\
teetotal	&0.00	&1.73	&1.73\\
equalization	&0.50	&1.64	&1.64\\
irresistibility	&1.57	&0.55	&1.57\\
teenage	&1.50	&0.00	&1.50\\
bikini	&1.38	&0.00	&1.38\\
navigation	&1.25	&0.50	&1.25\\
cardamom	&1.12	&0.25	&1.12\\
fertileness	&1.00	&0.00	&1.00\\
gainful	&0.86	&0.77	&0.86\\
climax	&0.75	&0.08	&0.75\\
collectable	&0.57	&0.62	&0.62\\
posthaste	&0.00	&0.50	&0.50\\
enamelware	&0.17	&0.42	&0.42\\
metaphoric	&0.33	&0.18	&0.33\\
directionality	&0.22	&0.00	&0.22\\
switchover	&0.12	&0.10	&0.12\\
appendix	&0.00	&0.00	&0\\
minuscule	&-0.14	&0.12	&-0.14\\
bobber	&0.12	&-0.42	&-0.42\\
miniaturization	&-0.62	&0.25	&-0.62\\
misrecognition	&-0.75	&-0.80	&-0.80\\
impel	&-1.00	&0.17	&-1.00\\
unselect	&-1.12	&-0.67	&-1.12\\
dodgers	&-0.89	&-1.20	&-1.20\\
nonplussed	&-1.29	&0.08	&-1.29\\
stifled	&-1.38	&-0.44	&-1.38\\
impractical	&-1.50	&-0.55	&-1.50\\
imperceptivity	&-1.56	&-0.56	&-1.56\\
varicella	&-1.62	&-0.20	&-1.62\\
prattler	&-1.67	&-0.92	&-1.67\\
gentrify	&-1.75	&-1.50	&-1.75\\
smoking	&-1.75	&-1.45	&-1.75\\
rant	&-1.86	&-1.55	&-1.86\\
defoliate	&-1.88	&-0.64	&-1.88\\
notoriously	&-1.89	&-1.55	&-1.89\\
debilitating	&-2.00	&-0.60	&-2.00\\
paralyze	&-2.00	&-0.67	&-2.00\\
pettiness	&-2.00	&-1.89	&-2.00\\
rift	&-2.00	&-1.09	&-2.00\\
bacteria	&-2.11	&-0.36	&-2.11\\
detractor	&-2.14	&-1.75	&-2.14\\
egocentrism	&-2.25	&-2.00	&-2.25\\
illegitimate	&-2.00	&-2.30	&-2.30\\
curse	&-2.43	&-1.75	&-2.43\\
slovenliness	&-2.38	&-2.55	&-2.55\\
inbreed	&-2.67	&-1.82	&-2.67\\
horrible	&-2.62	&-2.78	&-2.78\\
denigration	&-2.88	&-2.44	&-2.88\\
stalker	&-3.00	&-2.67	&-3.00\\
narcism	&-2.43	&-3.00	&-3.00\\

\hline
\end{tabular}
\caption{\label{tab:examples} {Randomly sampled terms and their anxiety-association score from Words of Warmth.}}
}
% \vspace*{-3mm}
% \end{center}
\end{table}

\subsection{FAQ}

\bl{Q1. When should one use the warmth score and when should one use trust and sociability scores?\\[3pt]
Ans. Decades of social cognition research has converged on two primary dimensions: competence and warmth. Thus, for many applications %exploring social cognition and stereotyes, 
it is useful to examine the warmth and competence dimensions (using corresponding lexicon entries).
More nuanced analysis is enabled by splitting the warmth dimension into sub-categories. This is particularly appropriate when trust and sociability are expected to diverge: e.g., modern-day politicians are often seen as untrustworthy, yet sociable. One may also use a specific sub-dimension (trust or sociability) if that is the focus of the work: e.g., if one is interested in exploring trust in AI-generated text towards targets of interest, then the trust lexicon can be used.\\[5pt]
Q2. What is the purpose of the popup feedback during the annotation process?\\[3pt]
Ans. Annotation can be a tedious process. So it is unfortunate when one misunderstands some directions and spends time producing a large number of poor annotations. The popup feedback is there to let annotators know (as they are annotating) when they get certain instances wrong so that they can assess whether they have misunderstood something. This way they get immediate feedback. Secondly, it helps with quality control---people tend to refrain from clicking randomly when they know these checks exist.

}

\subsection{AMT Questionnaires}
\begin{figure*}[t]
	     \centering
	     \includegraphics[width=0.85\textwidth]{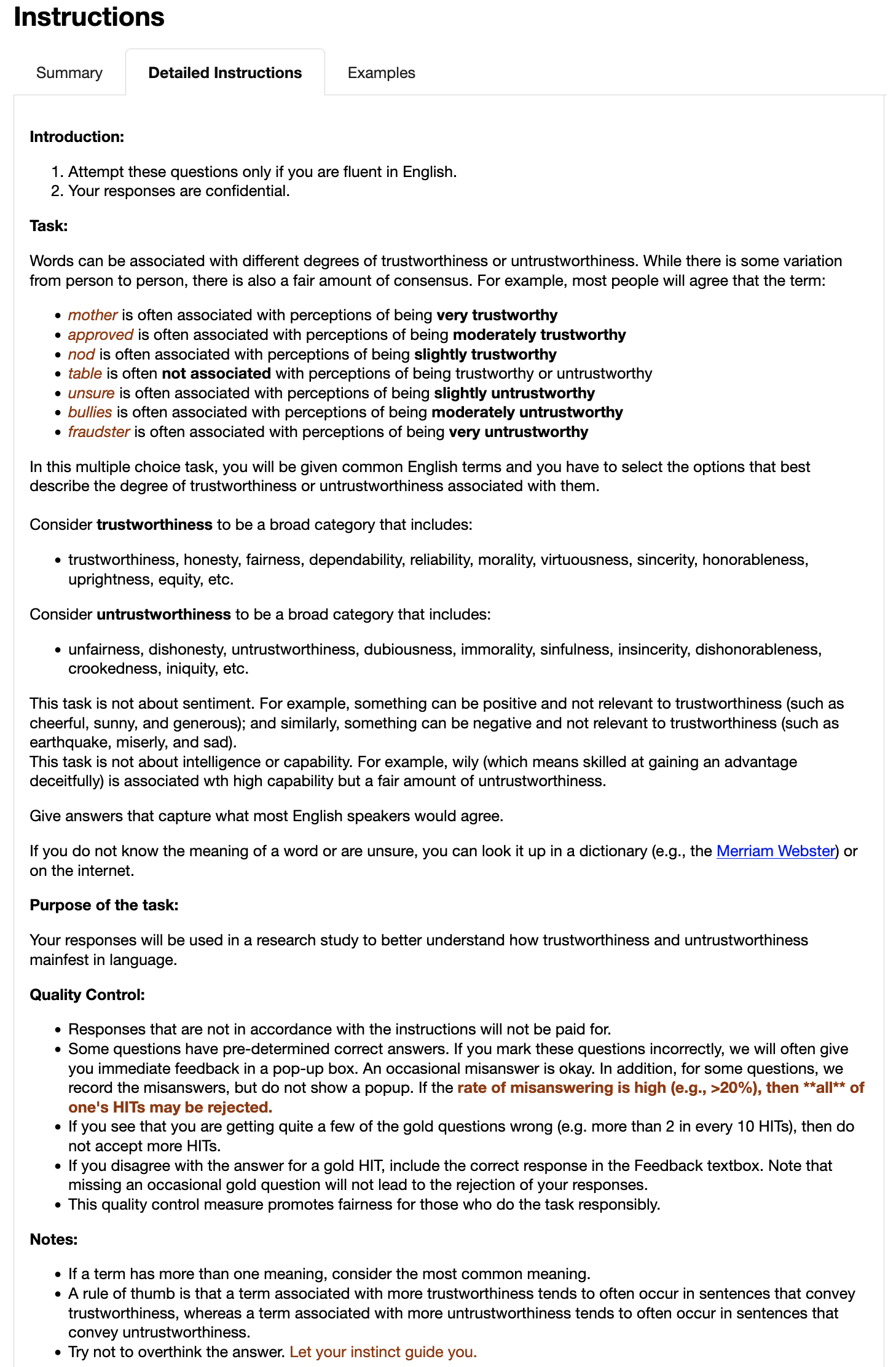}
	     \caption{Trust Questionnaire: Detailed instructions.}
	     \label{fig:trust-det-instr}

	 \end{figure*}

\begin{figure*}[t]
	     \centering
	     \includegraphics[width=\textwidth]{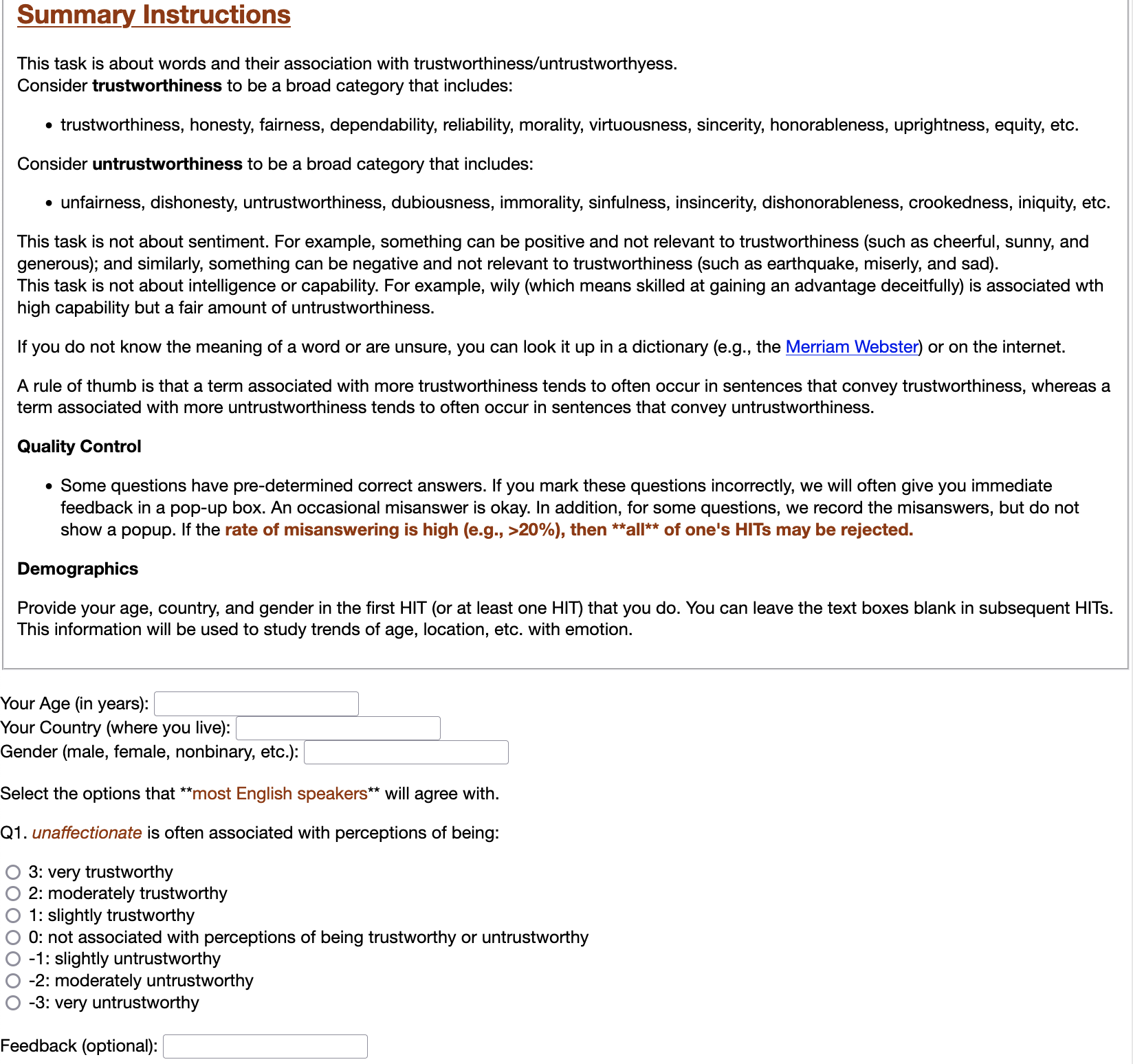}
	     \caption{Trust Questionnaire: Sample question.}
	     \label{fig:trust-q-main}
	 \end{figure*}

\begin{figure*}[t]
	     \centering
	     \includegraphics[width=0.85\textwidth]{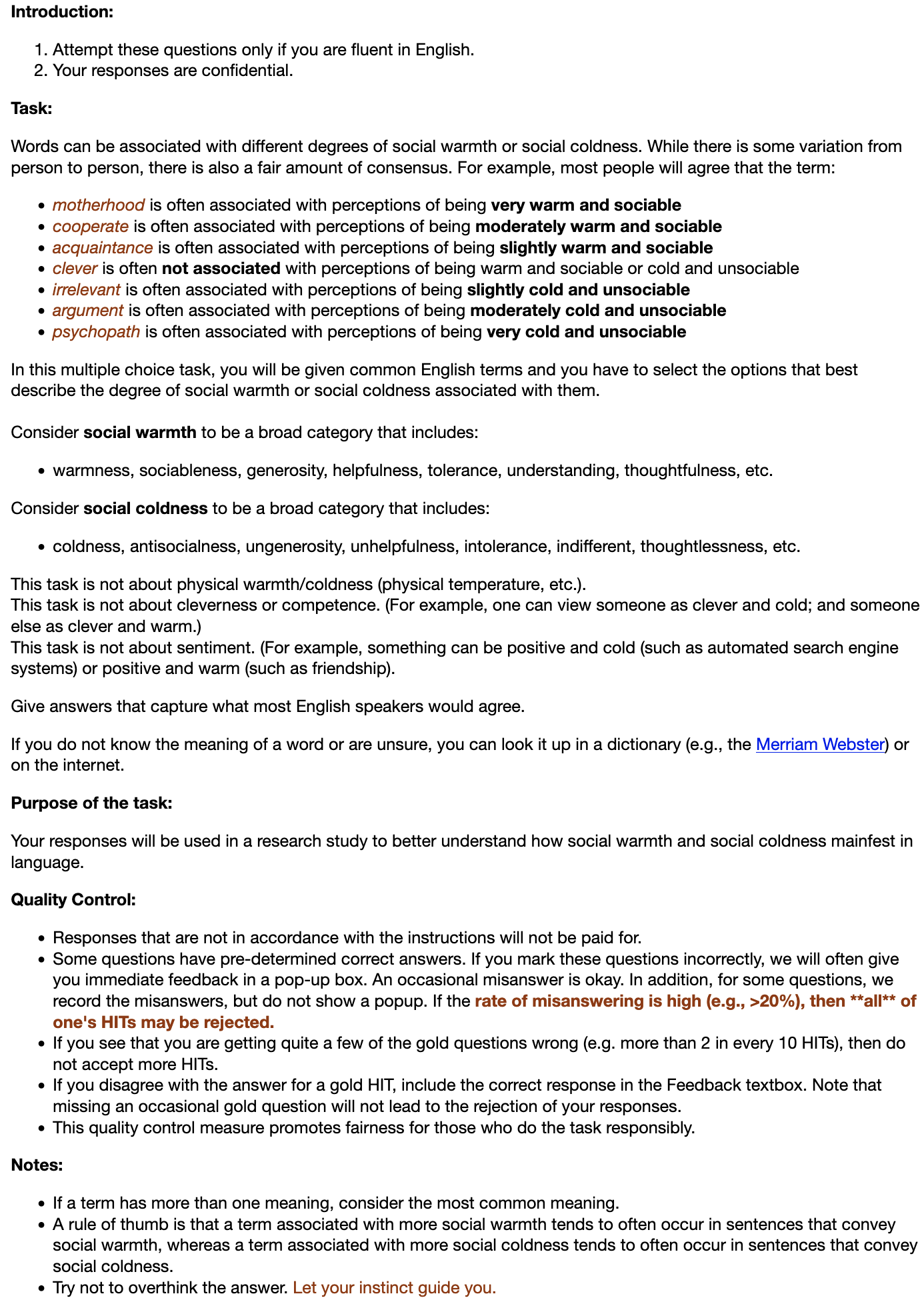}
	     \caption{Sociability Questionnaire: Detailed instructions.}
	     \label{fig:soc-det-instr}

	 \end{figure*}

\begin{figure*}[t]
	     \centering
	     \includegraphics[width=\textwidth]{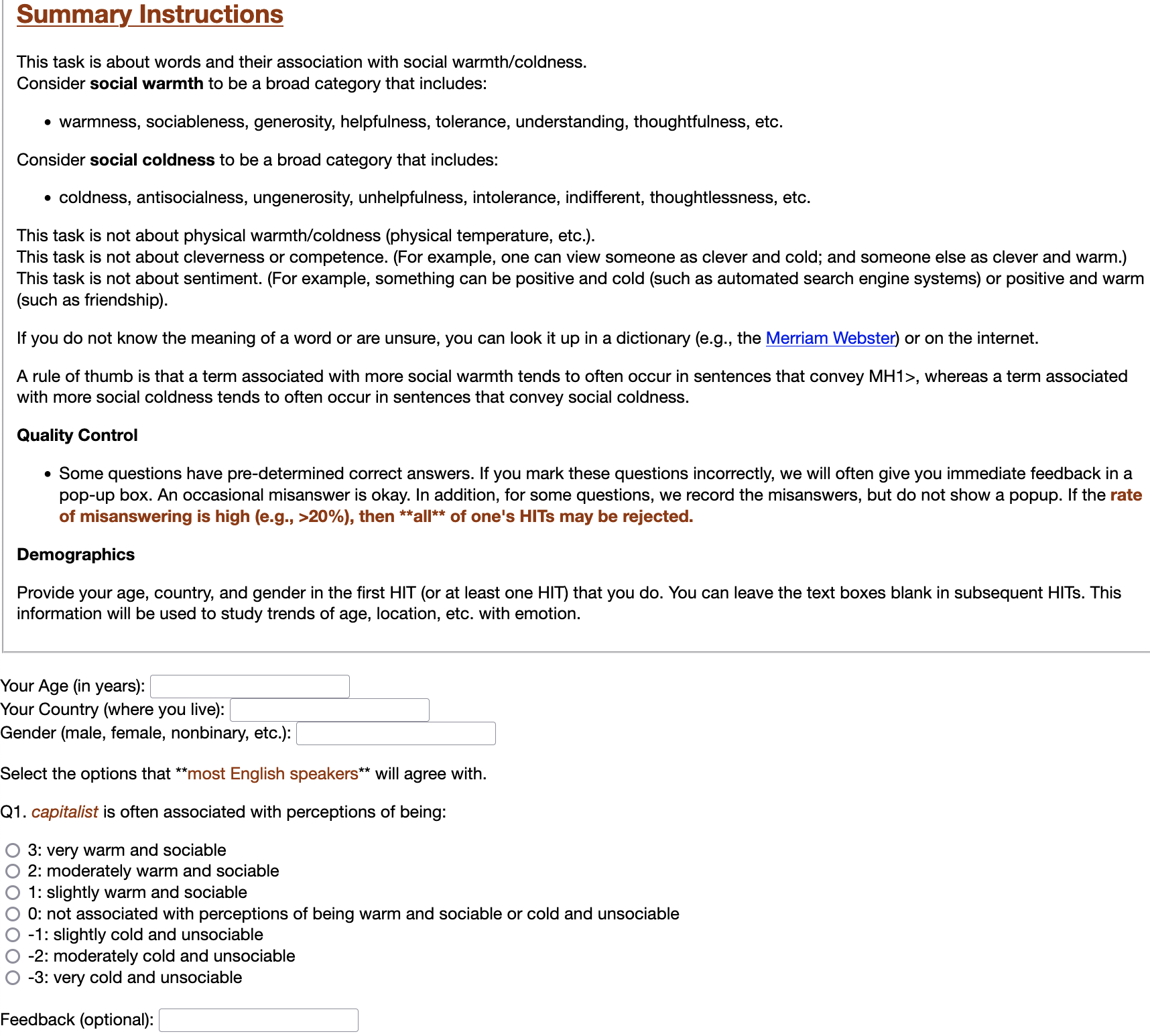}
	     \caption{Sociability Questionnaire: Sample question.}
	     \label{fig:soc-q-main}
	 \end{figure*}

\begin{figure*}[t]
	     \centering
	     \includegraphics[width=0.65\textwidth]{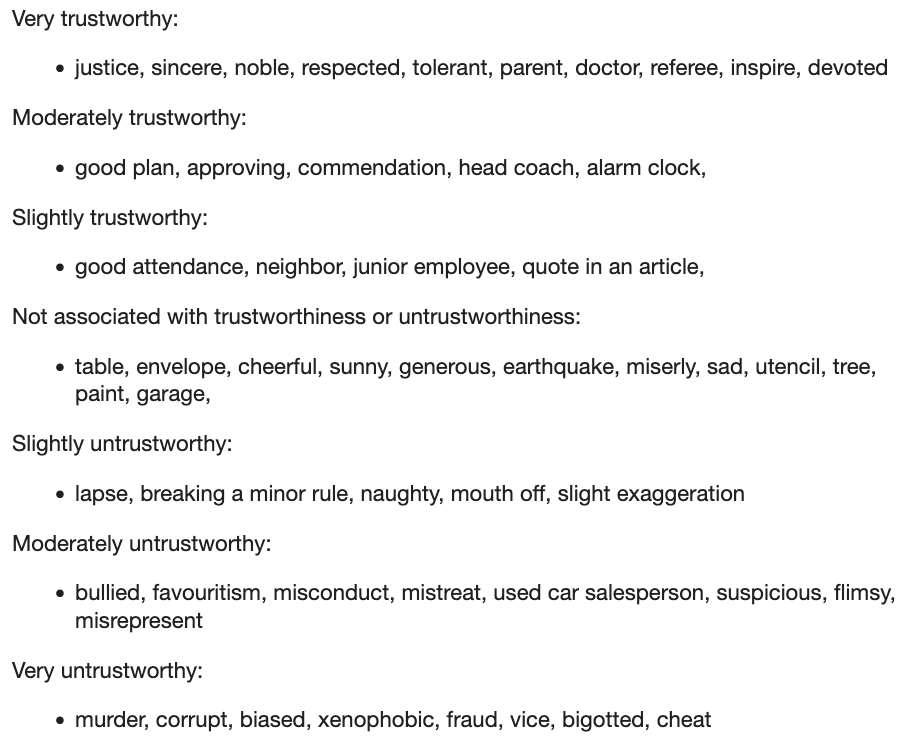}
	     \caption{Trust Questionnaire: Examples.}
	     \label{fig:trust-q-examples}
	 \end{figure*}

\begin{figure*}[t]
	     \centering
	     \includegraphics[width=0.65\textwidth]{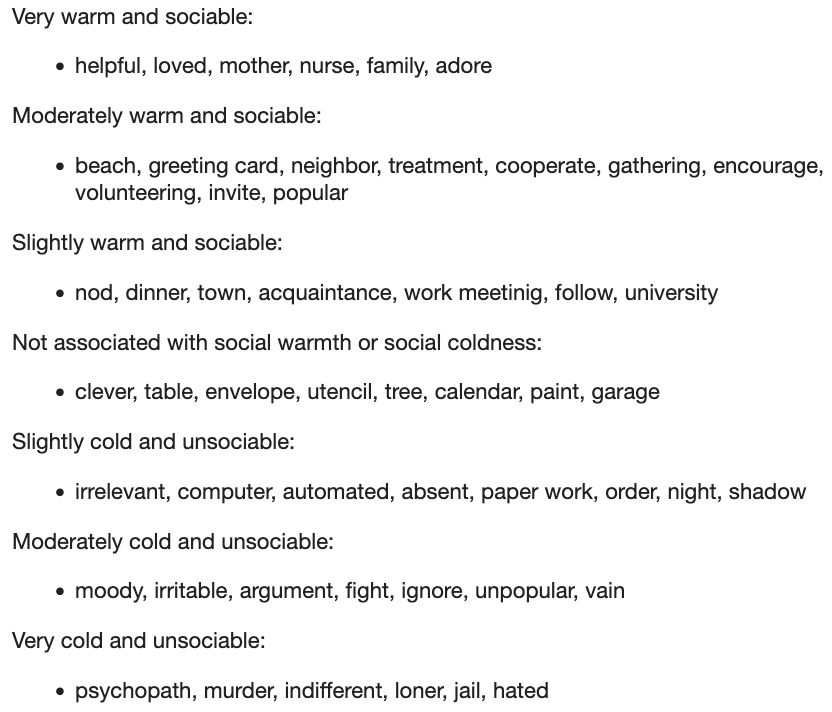}
	     \caption{Sociability Questionnaire: Examples.}
	     \label{fig:soc-q-examples}
	 \end{figure*}

Screenshots of the trust and sociability detailed instructions, sample question, and examples presented to the annotators are shown in Figures \ref{fig:trust-det-instr} through \ref{fig:soc-q-examples}. Participants were informed that they may work on as many instances as they wish. 

\subsection{Distribution of Words of Warmth}

Words of Warmth is made freely available on the project website as a compressed file. Terms of use will require that users not re-distribute the file and not post any form of the lexicon on the web. This is to prevent the resource being included in the data scrape fed to a large language model. 
See full list of terms of use at the project home page.
Table \ref{tab:examples} shows entries for a random sample of words from Words of Warmth.  

\clearpage

\newpage

%  \item SUBTLEX \cite{new2007use}:

% \item All 4,206 terms in the positive and negative lists of the General Inquirer \cite{Stone66}.\\[-19pt]
% \item All 1,061 terms listed in ANEW \cite{bradley1999affective}.\\[-19pt]
% \item All 13,915 terms listed in the \newcite{warriner2013norms} lexicon.\\[-19pt]
% \item 520 words from the Roget's Thesaurus categories corresponding to the eight basic Plutchik emotions.\footnote{http://www.gutenberg.org/ebooks/10681}\\[-19pt]
% Note that this set of terms includes both terms that are more common in social media communication (for example, {\it :), soannoyed, grrrrr, stfu}, and {\it thx})
% as well as regular English words.\footnote{Some of the terms included from the Twitter source were deliberate spelling variations of English words, for example, {\it bluddy} and {\it sux}.}\\[-22pt]

%  \begin{table}[t!]
% \begin{center}
% {\small
% \begin{tabular}{lclc}
% \hline
% \bf Word	&\bf Score$\uparrow$	&\bf Word	&\bf Score$\downarrow$ \\\hline
% \textit{infuriated}  & 3                     & \textit{applaud}         & -3\\
% \textit{traumatized} & 3                     & \textit{benedictional}   & -3\\
% \textit{warcrimes}   & 3                     & \textit{home}            & -3\\
% \textit{homicidal}   & 3                     & \textit{harmlessness}    & -3\\
% \textit{panicking}   & 3                     & \textit{melodiously}     & -3\\
% \hline
% \end{tabular}
% \caption{\label{tab:examples} {Example terms with the highest ($\uparrow$) and lowest ($\downarrow$) anxiety scores in Words of Warmth.}}
% }
% % \vspace*{-3mm}
% \end{center}
% \end{table}

  \clearpage

 \begin{figure}[t]
	     \centering
	     \includegraphics[width=0.5\textwidth]{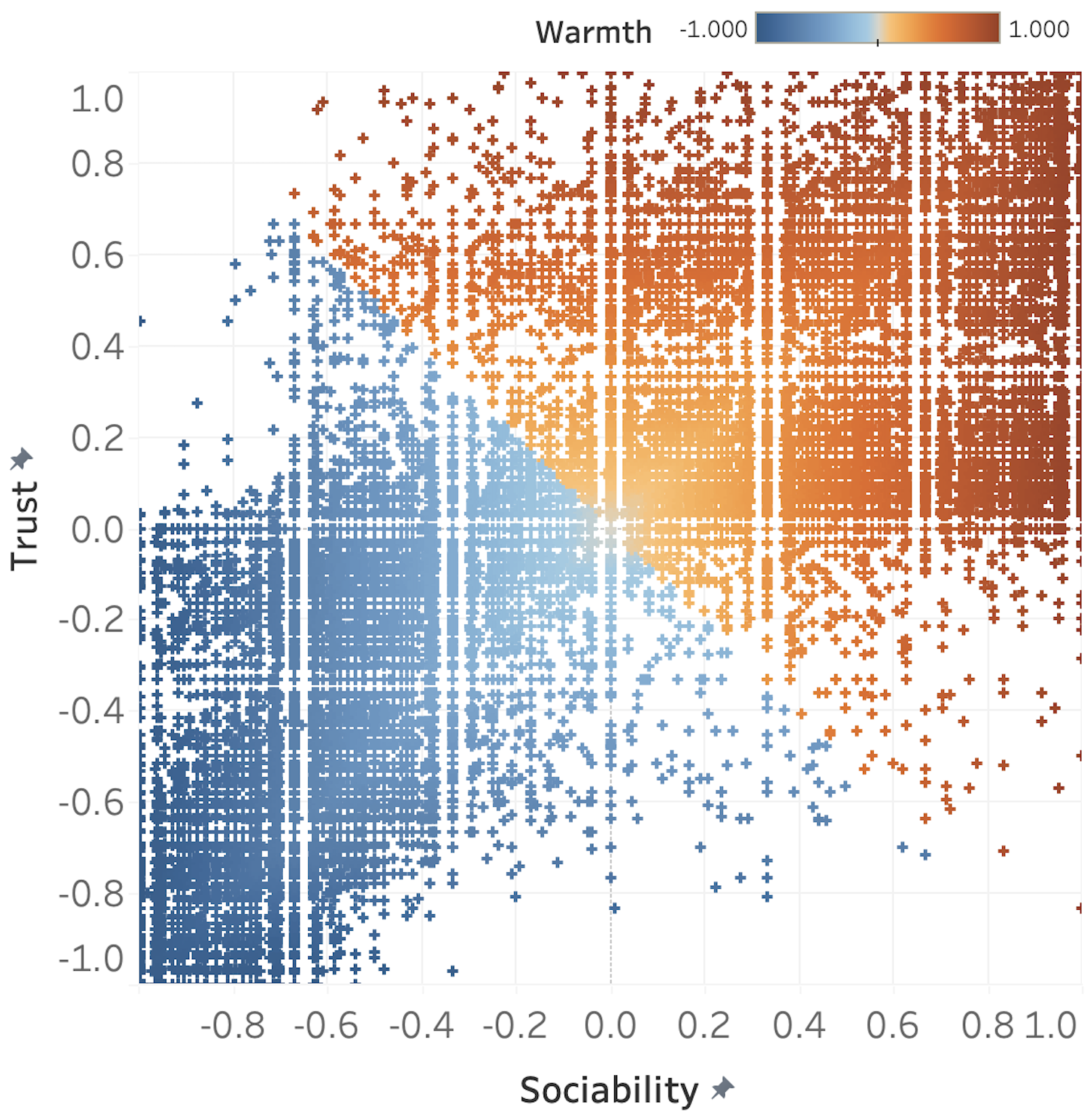}
	     \caption{Scatterplot of words on the  trust--sociability space. Individual points are colored as per their W score.}
	     \label{fig:WTS}

	 \end{figure}

\begin{figure}[t]
	     \centering
	     \includegraphics[width=0.5\textwidth]{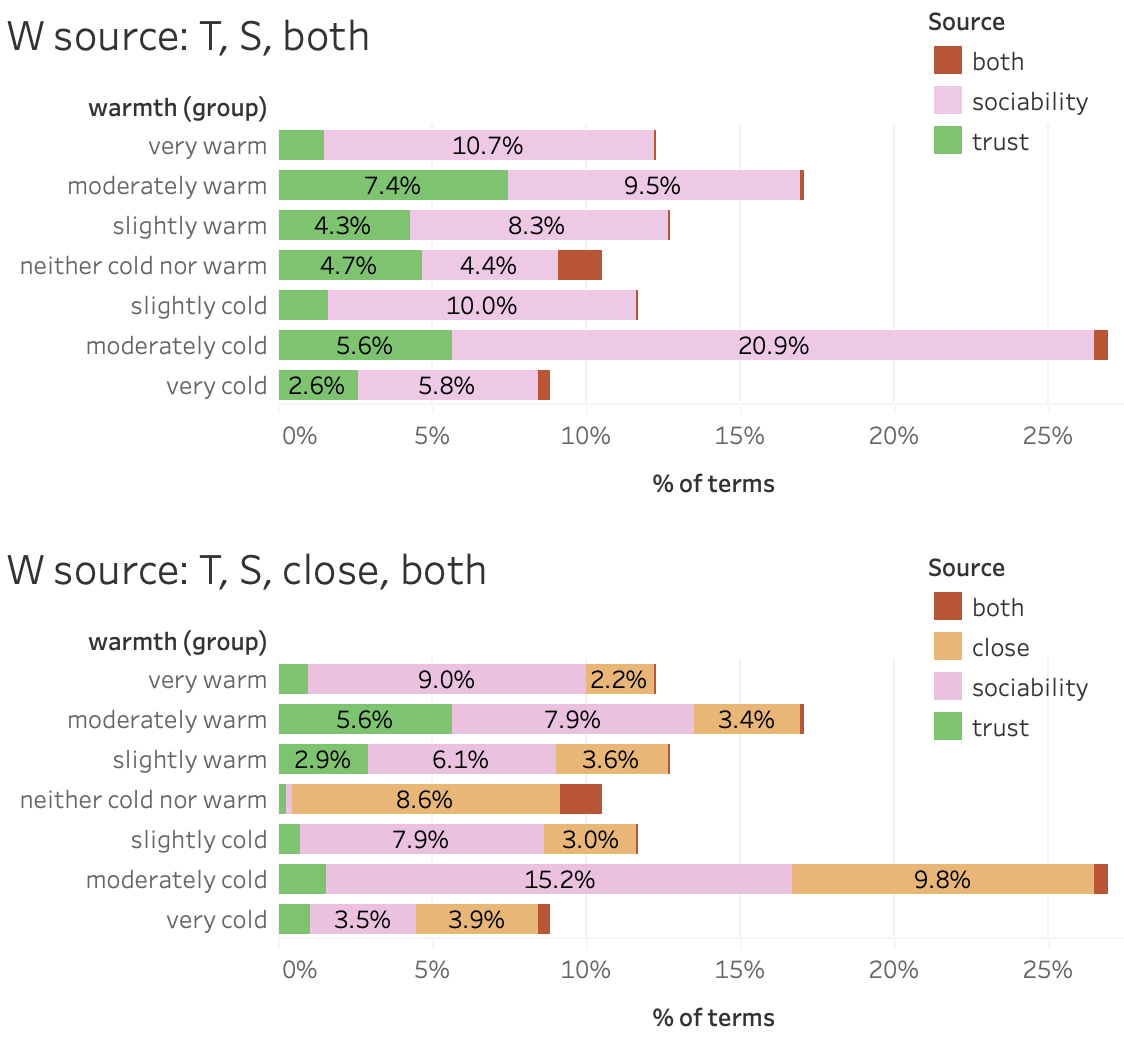}
	     \caption{Stacked bar charts showing 
         a break down of the percentage of terms in each of the warmth classes into the percentage of entries obtained from the trust lexicon and the percentage of entries obtained from the sociability lexicon.
         (a) shows a break down into 3 classes: percentage of terms for which the W score is the same as the T score and different from the S core (abs(T score) > abs(S score)), percentage of terms for which the W score is the same as the S score and different from the T score (abs(S score) > abs(T score)), and percentage of terms for which the W score is the same as both the T and S scores (abs(T score) = abs(S score)). (b) is similar to (a) except that a fourth class, \textit{close}, is added to show the cases where the S and T scores are close to each other (T score $-$ S score < 0.5).}
	     \label{fig:WbreakTS}

	 \end{figure}

\subsection{Case Study: Professions}

\begin{figure*}[t]
	     \centering
	     \includegraphics[width=\textwidth]{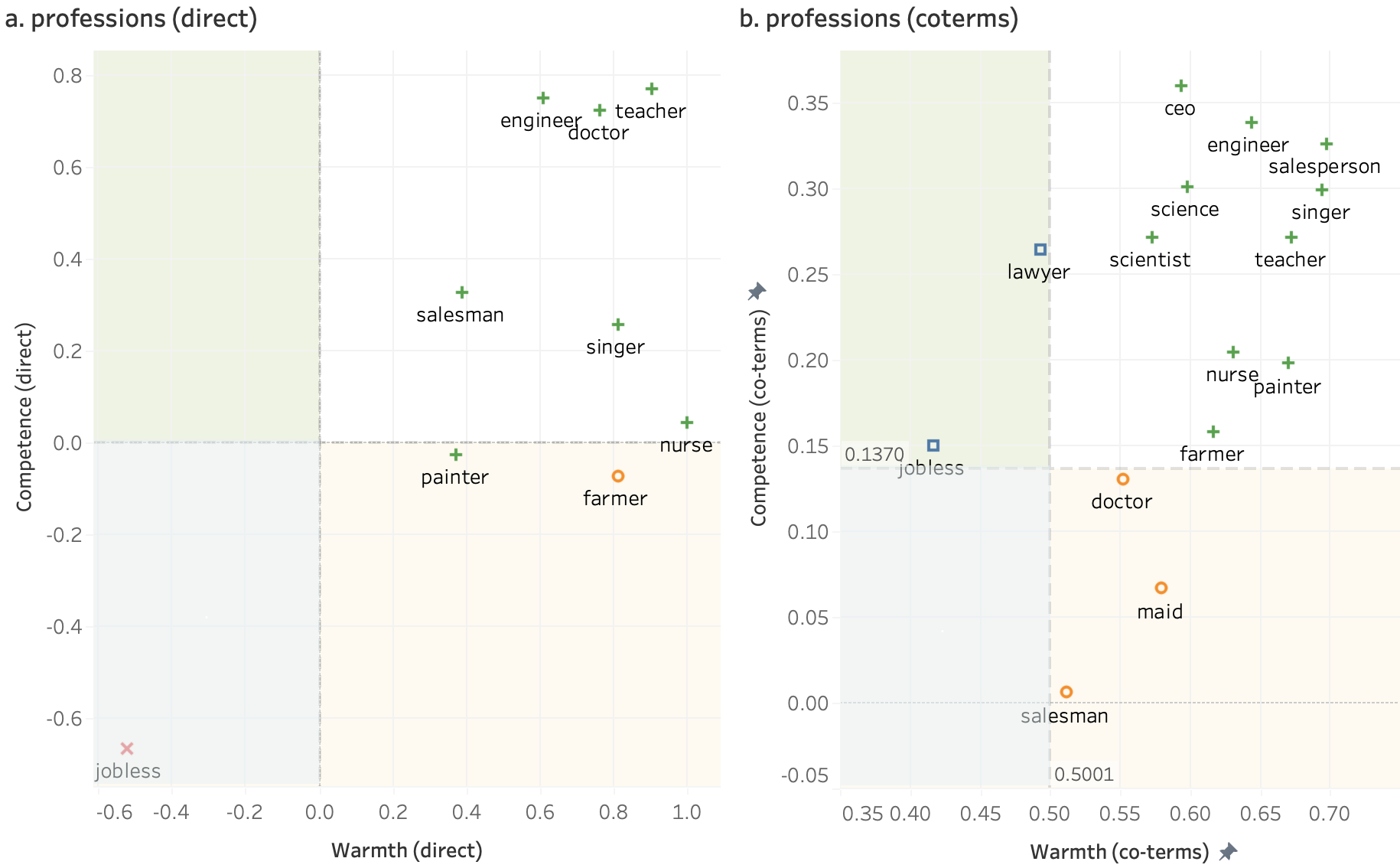}
      \caption{Direct and co-term W--C plots for various professions.}
	 
	     \label{fig:professions-wc}
      \vspace*{-3mm}
	 \end{figure*}

Figure \ref{fig:professions-wc} shows  W--C plots for various professions. The direct lexicon lookup of the targets (Fig \ref{fig:professions-wc} (a)) shown that people perceive engineers, doctors, and teachers to be high competence, whereas nurses and teachers are considered very warm. In contrast, being jobless is perceived as cold and incompetent. The coterms plot (b) shows that mentions of CEO get an even higher competence score than engineer and teacher, and in fact the mentions of doctor get a lower competence score than nurse.\footnote{The term CEO is not in the WC lexicons.} This indicates that even though doctors are considered as competent, their mentions in social medial are more in contexts where one is expressing a lack of competence/power/situational control (e.g., not having access to a doctor). 

\subsection{Supplementary Figures and Tables}

Figure \ref{fig:WTS} shows a scatter plot of words on the T--S space, colour coded as per their W score (described in Section 5). 
\bl{Figure \ref{fig:WbreakTS} shows a break down of the percentage of terms in each of the warmth classes into the percentage of entries obtained from the trust lexicon and the percentage of entries obtained from the sociability lexicon.}

\begin{figure*}[t]
	     \centering
	     \includegraphics[width=\textwidth]{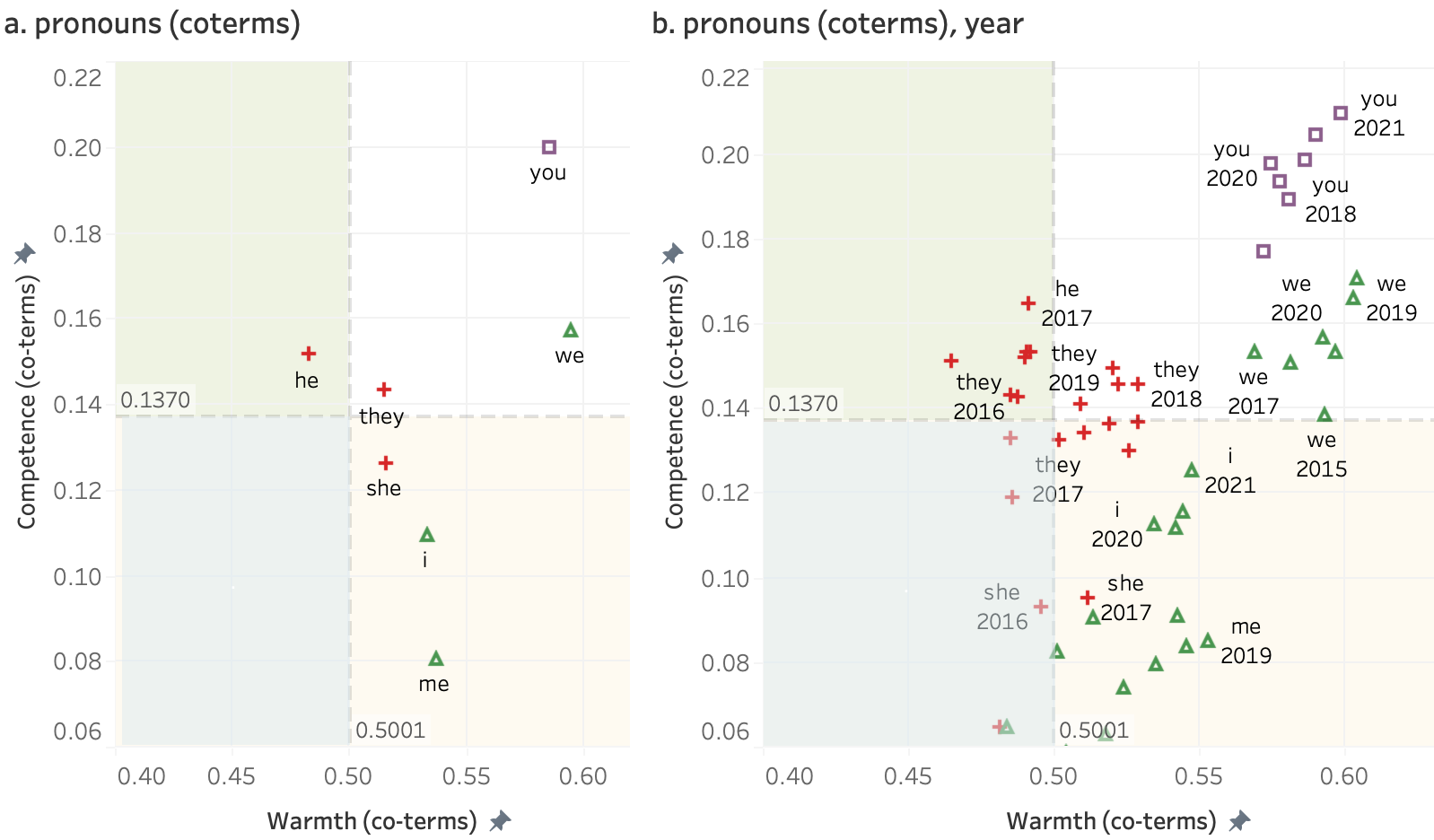}
      \caption{Direct and co-term W--C plots for various pronouns.}
	     \label{fig:pronouns-wc}
      \vspace*{-3mm}
	 \end{figure*}

Figure \ref{fig:pronouns-wc} (a) shows the W--C plot for various pronouns. Figure \ref{fig:pronouns-wc} (b) is the same plot except the pronouns are plotted separately for every year. (These plots were described in Section 7.)

% target-wc.twb

% target-wc.twb

\subsection{Computational Resources and Carbon Footprint}
A nice advantage of using simple lexicon-based approaches is the low carbon footprint and computational resources required. All of the experiments described in the paper were conducted on a regular personal laptop.

\end{document}